%% file: arXiv.tex
\documentclass[10pt,twocolumn]{article} 
\usepackage{simpleConference}
\usepackage{times}
\usepackage{graphicx}
\usepackage{amssymb}
\usepackage{url,hyperref}

\usepackage{graphicx}
\usepackage[caption=false]{subfig}
\usepackage{comment}
\usepackage{todonotes}
\usepackage{orcidlink}
\usepackage[square,numbers]{natbib}
\usepackage{graphicx}
\usepackage{amssymb}
\usepackage{booktabs}
\usepackage{nicefrac}
\usepackage{amsmath}
\usepackage{xspace}
\usepackage[normalem]{ulem}
\usepackage{arydshln}
\usepackage{bm}
\setuptodonotes{inline}
\usepackage{chessboard}

\usepackage{times}
\usepackage{siunitx}
\usepackage{adjustbox}

\newif\ifarxiv
\arxivtrue % or
\begin{document}
%
%\title{Can We Play Chess Efficiently with Vision Transformers?}
%\title{Applying Vision Transformers on Chess is not enough}
%\title{Vision Transformers Struggle to Play Chess}
%\title{Vision Transformers Struggle to\\ Master the Game of Chess}
%\title{ViT Struggles to Master the Game of Chess}
%\title{The Game of Chess Poses a\\ Challenge to Vision Transformers}
%\title{Representation Matters: The Game of Chess Poses a Challenge to Vision Transformers}
\title{Representation Matters for Mastering Chess: Improved Feature Representation in AlphaZero Outperforms Switching to Transformers}
%\title{Representation Matters:\\ The Game of Chess Challenges Transformers}

%\title{Vision Transformers Do not Master the Game of Chess (yet)}
% The Struggle of Mastering Chess with Vision Transformers

%Contribution Title\thanks{Supported by organization x.}}

\author{Johannes Czech$^{1}$\orcidlink{0000-0002-9568-8965} \quad
Jannis Blüml$^{1,2}$\orcidlink{0000-0002-9400-0946} \quad
Kristian Kersting$^{1,2,3,4}$\orcidlink{0000-0002-2873-9152}
\quad
Hedinn Steingrimsson$^{5,6}$ \orcidlink{0009-0008-3256-2995}
\\
\\
$^1$ Artificial Intelligence and Machine Learning Lab, TU Darmstadt, Germany \\
$^2$ Hessian Center for Artificial Intelligence (hessian.AI), Darmstadt, Germany \\
$^3$ Centre for Cognitive Science, TU Darmstadt, Germany \\
$^4$ German Research Center for Artificial Intelligence (DFKI), Darmstadt, Germany\\
$^5$ Department of Electrical and Computer Engineering, Rice University, Houston, TX, USA \\
$^6$ The Steingrimsson Foundation (safesystem2.org), Houston, TX, USA \\
\email{\small\{johannes.czech, blueml, kersting\}@cs.tu-darmstadt.de, \texttt{hedinn.steingrimsson@rice.edu}} \\
}

%
%\authorrunning{J. Czech et al.}
%\fi
%\titlerunning{Representation Matters: The Game of Chess Poses a Challenge to ViTs}

%
\maketitle              % typeset the header of the contribution

\newcommand{\AlphaZero}{\mbox{AlphaZero}\xspace}
\newcommand{\AlphaVile}{\mbox{AlphaVile}\xspace}

\begin{abstract}
While transformers have gained recognition as a versatile tool for artificial intelligence~(AI), an unexplored challenge arises in the context of chess --- a classical AI benchmark. Here, incorporating Vision Transformers (ViTs) into \AlphaZero is insufficient for chess mastery, mainly due to ViTs' computational limitations. The attempt to optimize their efficiency by combining MobileNet and NextViT  outperformed AlphaZero by about 30 Elo. However, we propose a practical improvement that involves a simple change in the input representation and value loss functions. As a result, we achieve a significant performance boost of up to 180 Elo points beyond what is currently achievable with \AlphaZero in chess.
In addition to these improvements, our experimental results using the Integrated Gradient technique confirm the effectiveness of the newly introduced features.
\end{abstract}

%%%%%
\input{content}

%%%%%

\vfill\null  % column break

\section*{Acknowledgements} 
The authors wish to express their gratitude to all those who have contributed to this study. The valuable insights, discussions, and constructive feedback provided by Hung-Zhe Lin and Felix Friedrich have significantly improved the quality of this work. The authors also extend their sincere thanks to Ofek Shochat, Lukas Helff and Cedric Derstroff for their thoughtful comments and suggestions, which have been instrumental in shaping the direction of this research. We are grateful for Daniel Monroe's insightful discussions.
%, and we also acknowledge the valuable contributions of chess grandmaster Hedinn Steingrimsson, whose expertise and discussions have improved the content of this paper.
We acknowledge the usage of ChatGPT, DeepL and Grammarly to enhance the language style of the paper.

\section*{Funding}
Funding for this research was partially provided by the Hessian Ministry of Science and the Arts (HMWK) through the cluster project ``The Third Wave of Artificial Intelligence - 3AI''.

\bibliographystyle{abbrvnat}
\bibliography{refs}

\clearpage
\appendix
\input{supplementary}

\end{document}

%% file: content.tex
\section{Introduction}

Transformers, a neural network architecture introduced in 2017 by Vaswani et al. ~\cite{vaswani2017attention}, have become one of the most dominant paradigms in modern Artificial Intelligence~(AI). Their range of applications has rapidly expanded, making them prevalent in tasks related to many areas of AI, including natural language processing, computer vision, and multimodal context learning. 
%Transformers have consistently demonstrated their capabilities by achieving groundbreaking performance, solidifying their position as a fundamental component of the AI model domain. 
By employing self-attention mechanisms, they differentiate themselves from traditional Convolutional Neural Networks (CNNs). This distinctive attribute allows the network to dynamically assess the significance of individual elements within the input sequence, providing an alternative to the limitations of sequential processing or fixed-size context windows. 
The rise in popularity of transformers can be attributed to their proficiency in handling long-range dependencies, a crucial characteristic of computer vision. 
%In this field, relationships between objects can be highly intricate and extend across vast spatial regions.
%Transformers are capable of capturing and modelling this information more effectively than CNNs, which may find it more challenging.
For this reason, transformer models are now favored over classical CNN approaches in various domains, including computer vision~\cite{dosovitskiy2020image}.

In the field of Reinforcement Learning~(RL), transformers hold great promise for creating robust models that can solve complex decision problems~\cite{chen2021decision,janner2021sequence}. These models can depict the connections and correlations between sequences of observations, actions, and rewards in the context of RL. They can be used to model the state representation, policy, and value function objectively~\cite{hu2023transforming,li2023survey}. In addition, they showcase superior performance as general world models~\cite{micheli2023transformers}.
\begin{figure*}[tp]
    \centering
    \includegraphics[width=\textwidth]{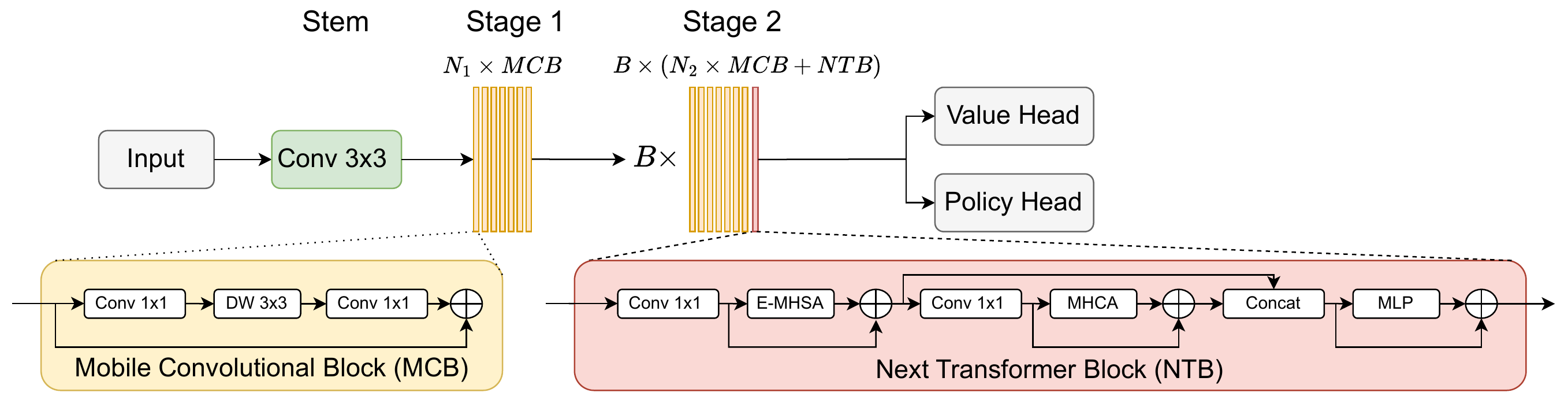}
    \caption{Architectural Overview of the Predictor Network in \AlphaVile. The Mobile Convolutional Block (MCB) is inspired by Sandler et al.'s work~\cite{sandler2018mobilenetv2}, while the Next Transformer Block (NTB) is integrated from Li et al.'s research~\cite{li2022next}. The parameter $B$ denotes the number of hybrid blocks within the architecture, offering scalability to the model. Our standard AlphaVile model employs ten MCBs in Stage 1 ($N_1 = 10$) and two Stage 2 Blocks ($B=2$). Each Stage 2 Block consists of seven MCBs ($N_2 = 7$) and one NTB.}
    \label{fig:final_rise_architectures}
\end{figure*}

The Transformer architecture is known for its impressive performance and versatility and has been compared to the ``Swiss Army Knife'' of AI.  However, the question remains: does it really live up to this analogy?
Relying on transformers solely due to their growing popularity across various domains does not necessarily lead to improvements. As demonstrated by Siebenborn et al.~\cite{siebenborn2022crucial}, the effect of transformer architecture on specific applications, including continuous control tasks, can result in differing outcomes. Their research found superior results by replacing the transformer with a Long Short-Term Memory network (LSTM). This emphasizes the nuanced nature of the transformer's suitability. 
While their capabilities are undeniable, the substantial scale of transformers, demanding billions of parameters for peak performance, imposes additional constraints. Substantial computational resources and memory are required, resulting in high latency and efficiency maintenance challenges~\cite{pope2022efficiently, xia2022trt}. These limitations become particularly significant in scenarios such as chess, where minimizing latency is crucial for computational efficiency.
This paper assesses the capabilities of transformers in playing chess, a benchmark game in the field of AI. A popular SOTA chess AI architecture is \AlphaZero, as introduced by Silver et al.~\cite{silver_mastering_2017, silver_general_2018}. The methodology seamlessly integrates neural networks with Monte-Carlo tree search (MCTS)~\cite{kocsis2006bandit}. 
%The outstanding achievements of \AlphaZero in games such as Go, chess, and shogi are widely recognized for surpassing earlier algorithms and even outstripping the highest-ranked human players.
\AlphaZero is appealing because it can learn from scratch, adapt to new challenges, and consistently perform well.
%These qualities make it a powerful model for machines to tackle complex problem-solving tasks.
In this paper, we present 
\AlphaVile\footnote{For more exhaustive information regarding our network architectures, input representations, and value loss formulations,
please check our supplementary material 
\ifarxiv
\else
on arXiv~\cite{czech2024representation}
\fi
and code on GitHub: \url{https://github.com/QueensGambit/CrazyAra/releases/tag/1.0.4}, accessed on 2023-10-26.},
%consult: \url{https://github.com/QueensGambit/CrazyAra/releases/tag/1.0.4}. Accessed on 2023-10-26.}
a novel convolutional transformer hybrid network. Incorporating a ViT within the \AlphaZero system enables testing of the potential of utilizing transformers and CNNs together to enhance chess performance.

%The code of our network architectures, input representations and different value loss formulations can be found in our repository
% State the contributions again
Although it is commonly believed that ``deep learning removes the need for feature engineering'', as argued by Francois Chollet in his book~\cite{chollet2021deep}, we believe that modifications to feature representation can generate improvements for methods such as \AlphaZero. It can help us in achieving similar goals as ``exploring the agent state space and having diverse agents with a heterogeneous skill set'', which can lead to "creative", broader and more diverse agent behavior~\cite{steingrimsson2021chess}.
Further, adding useful information to the features, i.e., the moves left in a chess game, can play a crucial role in making progress on an advanced chess task, outperforming much larger neural networks~\cite{steingrimsson2021chess}.
%The importance of the additional CNN head was discovered by introducing a new „System 2“ type dataset in  ~\cite{steingrimsson2021chess}, which put the logical reasoning and planning capabilities of SOTA chess architectures to the test and enabled taking a step towards developing Artificial General Intelligence.  
%~\cite{steingrimsson2021chess} was also the pioneering work that recommended a multi-agent approach „exploring the agent state space and having diverse agents with a heterogenous skill set”.  With a team of diverge agents, each conditioned by a latent variables which are designed to encourage the agents to play differently from one another, behavioral diversity was promoted. Also a strategy was laid out for the diversity to be achieved through intrinsic motivation, where each agent received a reward for exploring different parts of the state space, leading to variety of chess strategies and styles.

In addition to making architectural changes, we are investigating potential enhancements regarding the feature representation. We begin by introducing \AlphaVile and providing the necessary context. We then discuss the improved input representation and value loss, followed by presenting our empirical evaluation, which highlights the significant impact of our extended representation. Finally, we go over related work and draw a conclusion.

\section{\AlphaVile: Integrating transformers into AlphaZero}
\label{sec:alphavile}

AlphaZero, as introduced by Silver et al.~\cite{silver_mastering_2017, silver_general_2018}, represents a well-established model-based RL approach. Its primary strength is rooted in its ability to make predictions about the likely course of a given situation through the use of Monte-Carlo Tree Search (MCTS)~\cite{kocsis2006bandit} with an innovative integration of neural networks. 
Specifically, we make use of Prediction Upper Confidence bounds for Trees algorithm (PUCT) that was later refined in AlphaZero. For a detailed description of the PUCT algorithm, see ~\cite{steingrimsson2021chess}.

%In essence, AlphaZero excels in its capacity to evaluate the quality of potential actions and utilize this insight for forward planning, all while maintaining scalability.
%The effectiveness of \AlphaZero is mainly due to the innovative integration of search algorithms with neural networks

Our approach centers on the substitution of the residual network architecture (ResNet)~\cite{he2016deep}, a framework heavily reliant on convolutional layers, with a transformer-based architecture. Notably, elements such as the Monte-Carlo Tree Search (MCTS) algorithm and the loss function:
\begin{equation}
     \ell=\alpha\left[z-v\right]^{2}-\boldsymbol{\pi}^{\top} \log \boldsymbol{p}+c \cdot\|\theta\|_{2}^{2}\;,
\end{equation}
remain consistent with the original \AlphaZero design. Here, $[z-v]^{2}$ quantifies the mean squared error between the actual game outcome, denoted as $z$, and the predicted value $v$. Similarly, $\boldsymbol{\pi}^{\top} \log \boldsymbol{p}$ represents the cross-entropy between the target policy vector $\boldsymbol{\pi}$ and our predicted vector $\boldsymbol{p}$, a configuration adopted from Silver et al.~\cite{silver_mastering_2017}.
To further refine our model, we use the scalar parameter $\alpha$, serving as a weighting factor for the value loss. In our experiments, we set $\alpha$ to 0.01, a choice made to mitigate the risk of overfitting.

% add information on MCTS AlphaZero here
%\todo[inline]{TODO: Johannes rewrite from Dovovitskiy to ...Im summary AlphaVile... in such a way that first Figure 1 is explained which shows the AlphaVile architecture. Then the reasoning for choosing the different building blocks in AlphaVile is given with reference to prior work. Alternatively explaining Figure 1 and explaining what had been done before and what was selected as building blocks could be done together. Make it clear what was chosen in AlphaVile. Then can say ... In summary.
%One suggestion: As can be seen in Figure 1, the AlphaVile architecture consists of ... This is motivated by ...}
As can be seen in Figure~\ref{fig:final_rise_architectures}, \AlphaVile is the result of a synergistic fusion of components from AlphaZero~\cite{silver_mastering_2017}, NextViT~\cite{li2022next}, and MobileNet~\cite{howard2019searching}.
Our approach is based on the Next Hybrid Strategy as explained by Li et al.~\cite{li2022next}. In this strategy, a single transformer block is coupled with multiple convolutional blocks. The architectural configuration is further trimmed for optimal performance using TensorRT by combining different blocks and operations into a single block.
This is motivated by the work of Dosovitskiy et al.~\cite{dosovitskiy2020image} who conducted a comparative analysis that placed their vision ViT architecture in competition with SOTA ResNet~\cite{he2016deep} and EfficientNet~\cite{tan2019efficientnet} architectures, both reliant on CNNs. Their evaluation yields results that showcase ViT's superior performance in image classification tasks, particularly on well-established benchmark datasets such as ImageNet and CIFAR.
Subsequently, Han et al.~\cite{Han_2023} extend this exploration with another comprehensive evaluation, once again comparing ViT architectures to contemporary CNN-based counterparts. However, their study also highlights the concern of efficiency. Transformer models, by design, tend to be extensive and computationally more demanding than their CNN-based counterparts, often requiring extensive datasets for training. 
%Han et al. undertook an investigation into efficiency, aiming to explore whether a fusion of CNN and transformer architectures could yield improvements.
Han et al. emphasizes the symbiotic relationship that emerges when CNN and transformer models are combined.
Efforts are also being made to tackle the efficiency issues of ViTs, with a focus on improving performance. Innovations such as TensorRT as well as specially tailored architectures such as Trt-ViT~\cite{xia2022trt} and NextViT~\cite{li2022next} are contributing to the ongoing search for efficiency improvements.

In the context of CNNs, the MobileNet architecture was developed by Howard et al.~\cite{howard2019searching}. MobileNet innovatively combines depthwise separable convolutions with pointwise convolutions to reduce the computational overhead and memory demands typically associated with traditional CNNs, all while preserving high accuracy. MobileNets have consistently demonstrated their capability to achieve accuracy across a spectrum of computer vision tasks, all the while exhibiting significantly enhanced speed and memory efficiency compared to conventional CNNs. These characteristics make MobileNet a particularly well-suited choice for integration within the \AlphaZero framework. Various iterations of MobileNet have been introduced, including MobileNetV2 and MobileNetV3, each bringing additional optimizations and enhancements to the original architecture. For our work, we leverage the mobile convolution block, as presented in MobileNetV2 and reutilized in MobileNetV3.

Further, we employ stochastic depth techniques~\cite{huang2016deep}. This strategy serves a dual purpose, accelerating the training process while enhancing convergence. Additionally, we implement a scaling technique adapted from EfficientNet~\cite{tan2019efficientnet}, which facilitates the generation of networks in varying sizes to suit our needs. More details on this can be found in the supplementary materials 
\ifarxiv
section.
\else
section~\cite{czech2024representation}.
\fi
\begin{figure}[t]
    \centering
    \includegraphics[width=.9\linewidth]{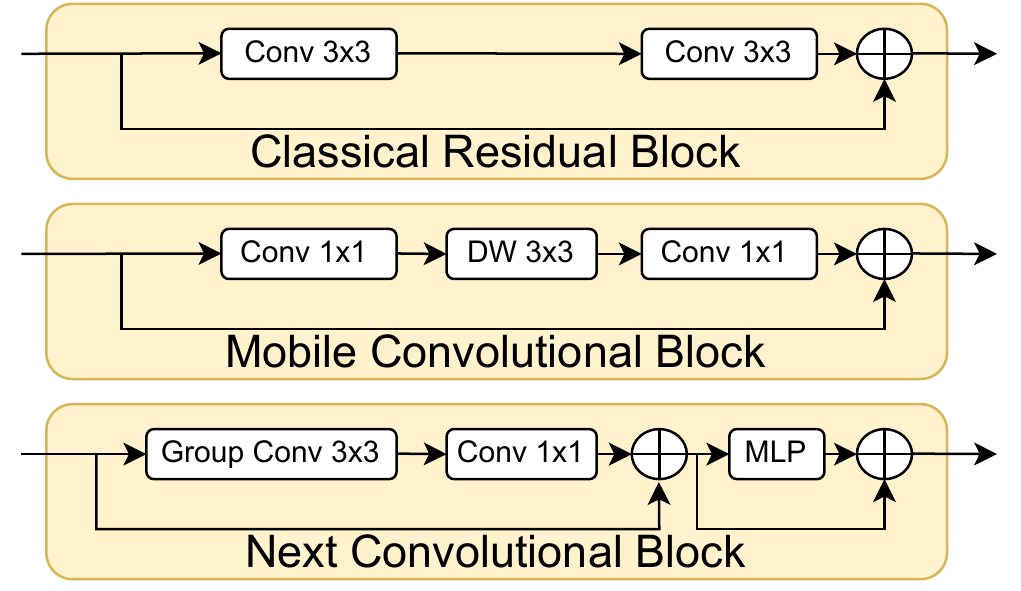}
    \caption{Comparing Architectural Components of Convolution-Based Blocks. This diagram utilises "DW" to denote Depthwise Convolution. Batchnorm and ReLU layers have been omitted for clarity. The conventional residual block, initiated by He et al.~\cite{he2016deep}, is substituted in \AlphaVile with the mobile convolution block, found in MobileNet, as stated by Sandler et al.~\cite{sandler2018mobilenetv2}. Additionally, we make use of the next convolution block originally introduced in NextViT by Li et al.~\cite{li2022next}}
    \label{fig:conv_blocks}
\end{figure}

To assess the performance of \AlphaVile, our evaluation begins with a comparative analysis of the MobileNet block compared to ResNet~\cite{he2016deep} and the ``Next Convolution Block'', a component that outperformed the \mbox{ConvNext} Transformer, PoolFormer, and Uniformer blocks in the study conducted by Li et al.~\cite{li2022next}.
These three convolutional blocks are visually represented in Figure~\ref{fig:conv_blocks}.
Notably, the mobile convolutional block~\cite{sandler2018mobilenetv2} demonstrates a slightly superior performance compared to the classical residual block~\cite{he2016deep}, while the next convolutional block~\cite{li2022next} exhibits notably inferior results under equivalent latency constraints. Consequently, based on experimental results in Table~\ref{tab:conv_blocks}, we select the mobile convolutional block as the default convolutional base block for \AlphaVile.

\begin{table*}[t]
\centering
\caption{Training Results for Comparing Convolutional Blocks in the Core of the Model. The mobile convolutional block, with an expansion ratio of three, outperforms the classical residual block and notably surpasses the next convolutional block. Furthermore, all three configurations exhibit similar latency on the GPU. The best results are highlighted in \textbf{bold}.}
\label{tab:conv_blocks}
\begin{tabular}{lcccccc}
\toprule
\textbf{Convolutional Block} & \textbf{Blocks} & \textbf{Channels} & \textbf{Combined Loss} & \textbf{Policy Acc. (\%)} & \textbf{Latency ($\mu s$)} \\
\midrule
Classical residual block~\cite{he2016deep} & 10 & 192 & 1.2350 $\pm$ 0.0031 & 57.50 $\pm$ 0.14 & 36.17 \\
Mobile conv. block~\cite{sandler2018mobilenetv2} & 9 & 256 & \textbf{1.2343} $\pm$ \textbf{0.0023} & \textbf{57.53} $\pm$ \textbf{0.05} & \textbf{34.78} \\
Next conv. block\cite{li2022next} & 10 & 256 & 1.2411 $\pm$ 0.0009 & 57.33 $\pm$  0.05 & 34.84 \\
\bottomrule
\end{tabular}
\end{table*}
Our investigation regarding the combination of convolutional base blocks with transformer blocks covers different sets of transformer blocks according to the integration strategy presented in Table~\ref{tab:transformer_blocks}. Following the proposal by Li et al.~\cite{li2022next}, we place a single transformer block after a given number of convolutional blocks, rather than grouping all transformer blocks at the end of the network.
For architectures with 15 convolutional blocks, our study shows that sparsity predominates. Smaller quantities of transformer blocks show better results compared to their more numerous counterparts. The configuration with precisely two transformer blocks is the optimal choice and yields the best results.
After introducing the \AlphaVile architecture, we now turn to the importance of representation.

\begin{table*}[t]
\centering
\caption{Grid Search Results for Different Transformer Block Configurations within the \AlphaVile Model, as Shown in Figure \ref{fig:final_rise_architectures}. Optimal performance, highlighted in \textbf{bold}, is achieved by including two level 2 blocks ($B$). The number of mobile blocks ($N_1, N_2$) is adjusted carefully to ensure the comparability of the models. The number of Transformer Blocks (NTBs) is set according to $B$.}
\label{tab:transformer_blocks}
\begin{tabular}{ccccccc}
\toprule
\textbf{\#Stage 2 Blocks (\bm{$B$})} & \bm{$N_1$} & \bm{$N_2$} & \textbf{Combined Loss} & \textbf{Policy Acc. (\%)} & \textbf{Latency ($\mu s$)} \\
\midrule
0 & 18 & 0 & 1.1901 $\pm$ 0.0049 & 58.67 $\pm$ 0.17 & 53.54 \\
1 & 8 & 8 & 1.1920 $\pm$ 0.0021 & 58.57 $\pm$ 0.17 & 54.40 \\
2 & 5 & 5 & \textbf{1.1887} $\pm$ \textbf{0.0065} & \textbf{58.67} $\pm$ \textbf{0.12} & 54.86 \\
3 & 5 & 4 & 1.2061 $\pm$ 0.0140 & 58.30 $\pm$ 0.43 & 54.67 \\
4 & 4 & 3 & 1.2327 $\pm$ 0.0045 & 57.57 $\pm$ 0.47 & \textbf{52.68} \\
\bottomrule
\end{tabular}
\end{table*}

\section{AlphaVile-FX: The importance of representation}
\label{sec:representation_matters}

In \AlphaZero, the traditional representation of the game state is a nuanced art. It manifests as a stack of planes, each of which encodes a particular facet of the complex state of the chessboard. These planes are structured as an $8 \times 8$ grid, with each cell serving as a single square on the chessboard.
Within this framework, there are two distinct plane types: \textit{bool} and \textit{int}. The first type describes planes where the value of each square is restricted to binary limits, i.\,e. 0 or 1. As an illustrative example, the first plane indicates the areas occupied by the first player's pawns, where the value 1 indicates their presence and 0 their absence. In contrast, the \textit{int} planes deal with integer numbers and provide a range of values instead of a binary contrast, e.\,g. the no-progress counter value is set on the entire 22nd plane. To improve computational robustness and numerical stability, these \textit{int} features are thoughtfully scaled to cover the floating point range from $-$1 to 1, using the extreme feature values as the reference points.
The original representation (Inputs V1.0) and can be found in Table~\ref{tab:input_representation}.
In total, Inputs V1.0 comprises 39 planes and gives our input a multidimensional structure: a tensor with dimensions $39\times8\times8$.
%and $52\times8\times8$ respectively.
% while Inputs V2.0 consists of 52 planes.
%It is worth noting that Inputs V1.0 was defined in the framework by Czech et al.~\cite{czech2021improving}.
%, thus consolidating its place in the contemporary chess AI landscape.

\input{tables/input_features_v2}

%\subsection{Extending the representation}

%"In the pioneering work ~\cite{steingrimsson2021chess}, the importance of feature representations was proven by performing rigorous experiments on a new advanced "System 2" type chess task, that SOTA chess architectures still struggle with. There the best performing architecture, which included just like in our experiments, both a moves-left and WDL heads, outperformed significantly larger neural networks. Furthermore, in ~\cite{steingrimsson2021chess} the importance of behavioral diversity achieved by „exploring the agent state space and having diverse agents with a heterogenous skill set” resulting in different agents that play differently from one another was first pointed out. By giving each agent a reward for exploring different parts of the state space, a variety of chess strategies and styles which could be described as “creativity” could be achieved. Our approach where we gave our agent access to broader information was motivated by this approach."

\subsection{Expanding the input representation}
It is a generally accepted principle that the role of representation is central to traditional machine learning, but there is an ongoing discussion about its importance in the field of deep learning.
%With this in mind, we begin our investigation by asking: How can we improve the representation of our input features and loss function, regardless of the architecture available to us?
In this paper, we present a novel interpretation, as shown in both the top and bottom segments of Table~\ref{tab:inputs_v2}, of the input by introducing additional features into the existing framework while also removing two features. 
%To prevent overfitting and improve prediction quality, the extensions for this task include representing the material disposition of both players and excluding color information and the current number of moves. 
We exclude the color information as this can distort the evaluation in favor of the White player regardless of the underlying position, as White has often an advantage in chess positions. The differentiation between the active player to move and the opponent player still persists. Moreover, we remove the current move number, but not the no-progress counter for the 50 move rule.
Additionally, we add several features including two masks for all pieces of each player, a checkerboard pattern, the relative material difference, a boolean map signaling if there are opposite color bishops, all checking pieces and the overall material count of the current player.
% maybe cite the paper here which goes over exploits in chess?
The new input definition brings significant improvements, particularly regarding policy and value loss functions. We argue that these supplementary features, though derivable from existing features, enhance the network's capacity by providing essential information in advance, thereby eliminating the need for in-network computations.
Table~\ref{tab:inputs} presents evidence that highlights the significance of feature engineering, leading to an advantage of about 100 Elo and demonstrating its continued relevance in the realm of deep neural networks.

\subsection{Redefining the value loss representation}
\label{sec:wdlp}
In order to boost the performance of chess engines, it is necessary to improve the quality of chess engines' ability to evaluate a given position. Taking inspiration from ~\cite{steingrimsson2021chess}, where it was shown that the approach led to performance improvements on a dataset called chess fortresses, we explore an inventive method advocated by Henrik Forstén, embodied by the Win-Draw-Loss-Head (WDL) framework\footnote{Further information can be found at \url{https://github.com/LeelaChessZero/lc0/pull/635}, accessed on 2022-11-11.}. This framework accurately predicts the percentage distribution of winning, drawing, and losing scenarios, while also introducing the Moves Left Head\footnote{Further details can be accessed at \url{https://github.com/LeelaChessZero/lc0/pull/961}, accessed on 2022-11-11.} to forecast the remaining number of moves until the game's completion. 
To achieve this, we have incorporated an additional output into the value head. This enhanced model, dubbed the WDLP Value Head, accurately predicts the number of half moves left until the conclusive end of the game. 
%To this end, we have introduced a new component in the value head called the WDLP Value Head. This improved model also predicts the number of half moves (plies) remaining until the end of the game.
Originally developed for finishing off won endgames, \citet{steingrimsson2021chess} discovered that the WDL approach is suitable more generally for complex chess tasks.

The following equation derives the classic value output~($v$). The parameter is bounded within the interval $\left[-1, +1\right]$ and derived from the  interplay of $\text{L}_{output}$ and $\text{W}_{output}$:
\begin{equation}
v = - \text{L}_{output}+ \text{W}_{output} \quad .
\end{equation}

In this formulation, $\text{L}_{output}$ indicates the likelihood of experiencing a defeat in the game, while $\text{W}_{output}$ depicts the probability of achieving victory.
%This alignment of variables presents a more detailed view of the game's inherent dynamics. 
To validate our findings empirically, we refer to the information extracted from Table~\ref{tab:wdlp}. The results clearly demonstrate the advantages of this novel weight loss approach, leading to a 33 Elo improvement.

In this redefined framework, we employ a consistent value policy loss, which is accompanied by an auxiliary goal of the remaining number of plies. This results in a significantly transformed loss function
\begin{equation}
    \ell = -\alpha (\mathbf{WDL}^{\top}_\text{t} \log \mathbf{WDL}_\text{p}) - \boldsymbol{\pi}^{\top} \log \boldsymbol{p} + \beta (ply_\text{t} - ply_\text{p})^2 + c \|\theta\|_{2}^{2}
\end{equation}
employing $\mathbf{\mathbf{WDL}}_\text{p}$, a probability distribution that predicts the probabilities of win, draw, or loss, while $\mathbf{\mathbf{WDL}}_\text{t}$ defines the target distribution. 
Within this framework, the scalar parameters $\alpha$ and $\beta$ allow weighting each loss component.

\begin{table*}[t]
\centering
\caption{Understanding the Impact of Input Representations on Performance Metrics.
This table presents experimental results from different input representations, highlighting their impact on value and policy loss. The adapted Inputs V.2.0 is shown to be the most sophisticated, signaling the ongoing quest for optimization.}

\label{tab:inputs}
\begin{tabular}{ccccccc}
\toprule
\textbf{Input Representation} & \textbf{Combined Loss} & \textbf{Policy Acc. (\%)} & \textbf{Value Loss} & \textbf{Latency ($\mu s$)} & \textbf{Elo Difference}\\
\midrule
%Inputs V.1.0    & 18    & 205      & 1.18773 $\pm$ 0.00399    & 58.733 $\pm$ 0.124 & 0.4442 $\pm$ 0.0013       & 19,202       \\
Inputs V.1.0 & 1.1918 $\pm$ 0.0028 & 58.63 $\pm$ 0.05 & 0.4448 $\pm$ 0.0007 & \textbf{52.08} & -\\
Inputs V.2.0 & \textbf{1.1901} $\pm$ \textbf{0.0049} & \textbf{58.67} $\pm$ \textbf{0.17} & \textbf{0.4371} $\pm$ \textbf{0.0002} & 53.54 & 96.7 $\pm$ 30.4\\
\bottomrule
\end{tabular}

\end{table*}

\begin{table*}[t]
\centering
\caption{Finding the Optimal Value Head for Chess Engines. This table reveals the results of two value head types. The Win-Draw-Loss-Ply (WDLP) value head emerges as the winner.}
\label{tab:wdlp}
\begin{tabular}{cccccc}
\toprule
\textbf{Value Head Type} & \textbf{Combined Loss} & \textbf{Policy Acc. (\%)} & \textbf{Value Loss} & \textbf{Latency ($\mu s$)} & \textbf{Elo Difference}\\
\midrule
MSE & 1.1933 $\pm$ 0.0021 & 58.50 $\pm$ 0.08 & 0.4406 $\pm$ 0.0002 & \textbf{53.35} &  -\\
WDLP & \textbf{1.1901} $\pm$ \textbf{0.0051} & \textbf{58.73} $\pm$ \textbf{0.12} & \textbf{0.4356} $\pm$ \textbf{0.0006} & 53.38 & 33.2 $\pm$ 19.0\\
%WDLP with MLP & 18 & 205 & 1.18923 $\pm$ 0.00228 & 58.600 $\pm$ 0.141 & 0.4381 $\pm$ 0.0007 & TODO \\
%WDLP with MLP & 18 & 205 & 1.19009 $\pm$ 0.00489 & 58.667 $\pm$ 0.170 & 0.4371 $\pm$ 0.0002 & 18,679 \\
\bottomrule
\end{tabular}
\end{table*}

Our experiments reveal that the incorporation of input features and optimization of the loss function enhance performance, leading to an improvement in both loss and accuracy. Therefore, we introduce the "FX" suffix to indicate our \textbf{F}eature e\textbf{X}tension --- a combination of the expanded input representation and the WDLP value head.
% These results demonstrate our ongoing commitment to push the boundaries of artificial intelligence.

\section{Investigating the significance of representation in chess mastery}

This section presents an empirical study of the \AlphaZero, \AlphaVile models and their \mbox{``-FX''} variants. We test these across the dimensions of accuracy, latency, and overall playing strength while providing comprehensive context through comparative assessments against other baseline models.
Each experimental investigation adheres to a carefully defined set of training hyperparameters, as detailed in the supplementary
\ifarxiv
materials.
\else
materials~\cite{czech2024representation}.
\fi
%Table~\ref{tab:hyperparams}.
Our research efforts are strengthened by using three different seeds in our different configurations.
%, thereby increasing the robustness and statistical significance of our results.
We utilize the KingBase Lite 2019 dataset\footnote{\url{https://archive.org/details/KingBaseLite2019}, accessed on 2022-11-02} as for training chess networks. This extensive collection of chess data comprises of more than one million games played by expert human players since 2000, each with an Elo rating of over 2200.
Furthermore, our scientific exploration includes an ablation study in the field of alternative chess variations, in particular atomic and crazyhouse.  Utilizing the lichess.org variant database\footnote{\url{https://database.lichess.org/\#variant_games}, accessed on 2023-10-23}, we extract and examine data from the top decile of players. This approach provides a valuable perspective on model performance in diverse and challenging chess scenarios.
The latency evaluation is performed on a NVIDIA GeForce RTX 2070 OC, using a batch size of 64 and benefiting from the advanced TensorRT-8.0.1.6 backend for accelerating throughput.

\subsection{Trade-off between efficiency and accuracy}

\begin{comment}
\begin{figure} [t]
    \centering
     \includegraphics[width=.65\linewidth]{media/nn_comparision/vit_move_acc_comparision_chess.pdf}
    \caption{Comparison among \AlphaVile and efficient networks with respect to accuracy-latency trade-off. Results are measured using three seeds. Details can be found in Table~\ref{tab:final_performance}}
    \label{fig:final_performance}
\end{figure}
\end{comment}

\begin{figure}[t]
    \centering
    \begin{minipage}{.49\textwidth}
        \centering
        \includegraphics[width=0.8\linewidth]{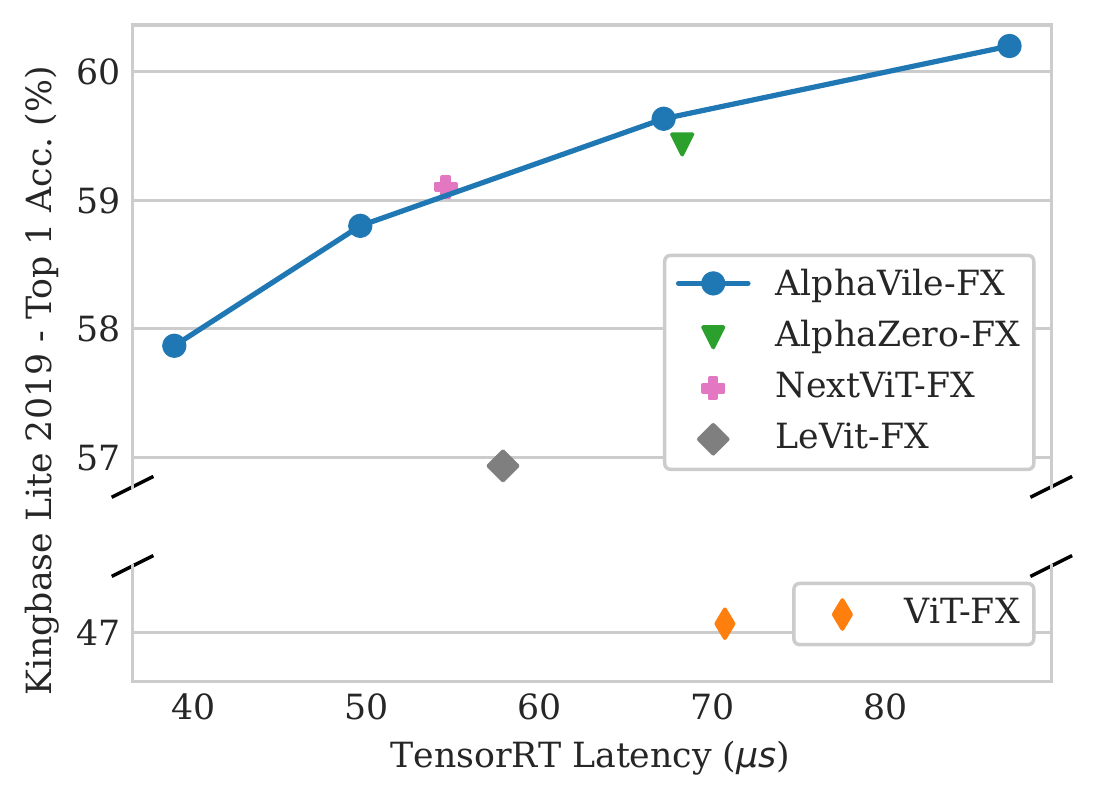}
    \caption{A comparison between \AlphaVile and other efficient neural network architectures, with a focus on achieving an optimal balance between accuracy and latency. The results were obtained from three independent seed runs.}% Table~\ref{tab:final_performance} provides a detailed report on the performance metrics.}
        \label{fig:final_performance}
    \end{minipage}%
    \hfill
\end{figure}

As Han et al.~\cite{Han_2023} demonstrate that the integration of transformers is widely praised for its versatility, but also presents inherent latency concerns. This is particularly noticeable in competitive contexts where precision and swift processing of large datasets are of importance. To decrease latency without sacrificing accuracy, we suggest combining convolutional base blocks and transformer blocks in our architecture.
We present four different configurations of our \AlphaVile architecture to illustrate the impact of network size on latency, as detailed in Table~\ref{tab:network_sizes}.
%As the \AlphaVile architecture, cf. Section~\ref{sec:alphavile} is petite with only 17 blocks ($N_1 = N_2 = 5, B = 2$) and has low latency, we have made adjustments to align it with latency levels similar to the other architectures.

\ifarxiv
\begin{table}[t]
    \centering
    \caption{Architectural Configurations of \AlphaVile in Different Sizes. Note: All versions feature a channel expansion ratio of 2 and use a combination of 50\,\% 3$\times$3 and 50\,\% 5$\times$5 convolutions. We use base channel counts that are dividable by 32 for faster inference.}
    \resizebox{\columnwidth}{!}{ % Adjust to fit the entire table to the text width
    \begin{tabular}{lccccc}
        \toprule
        \textbf{Size} & \bm{$B$} & \bm{$N_1$} & \bm{$N_2$} & \textbf{\# Blocks} & \textbf{Base Channels}\\
        \midrule
        AlphaVile (tiny) & 1 & 8 & 6 & 15 & 192 \\ 
        AlphaVile (small) & 1 & 11 & 10 & 22 & 192 \\
        AlphaVile (normal) & 2 & 10 & 7 & 26  & 224\\
        AlphaVile (large) & 2 & 13 & 11 & 37 & 224 \\
        \bottomrule
    \end{tabular}
    } % End of \resizebox
    \label{tab:network_sizes}
\end{table}
\else
\begin{table}[t]
    \centering
    \caption{Architectural Configurations of \AlphaVile in Different Sizes. Note: All versions feature a channel expansion ratio of 2 and use a combination of 50\,\% 3$\times$3 and 50\,\% 5$\times$5 convolutions. We use base channel counts that are dividable by 32 for faster inference.}
    \begin{tabular}{lccccc}
        \toprule
        \textbf{Size} & \bm{$B$} & \bm{$N_1$} & \bm{$N_2$} & \textbf{\# Blocks} & \textbf{Base Channels}\\
        \midrule
        AlphaVile (tiny) & 1 & 8 & 6 & 15 & 192 \\ 
        AlphaVile (small) & 1 & 11 & 10 & 22 & 192 \\
        AlphaVile (normal) & 2 & 10 & 7 & 26  & 224\\
        AlphaVile (large) & 2 & 13 & 11 & 37 & 224 \\
        \bottomrule
    \end{tabular}
    \label{tab:network_sizes}
\end{table}
\fi

We begin our investigation by examining the performance of a fully transformer-based neural network, the ViT \cite{dosovitskiy2020image}, when integrated with \AlphaZero. In order to ensure a fair comparison of latency, we trimmed these ViT models to match the latency of our \AlphaVile architecture, as employing a ViT model with an equivalent number of blocks would considerably increase latency. A comparison of these networks is and depicted in Figure~\ref{fig:final_performance}. These networks display a relatively high loss and decreased accuracy when compared to \AlphaZero.
Consequently, we start incorporating convolutional and transformer blocks. This assessment incorporates LeViT~\cite{graham2021levit}, NextViT~\cite{li2022next}, and our proposed \AlphaVile design. These strategies offer a better solution than the ViT-based approach and deliver better outcomes than \AlphaZero.

\ifarxiv
\newcommand\figwidth{0.8}
\else
\newcommand\figwidth{0.84}
\fi

\begin{figure}[ht!]
\centering
\subfloat[Performance evaluation in chess.]{
    %\centering
    \includegraphics[width=\figwidth\linewidth]{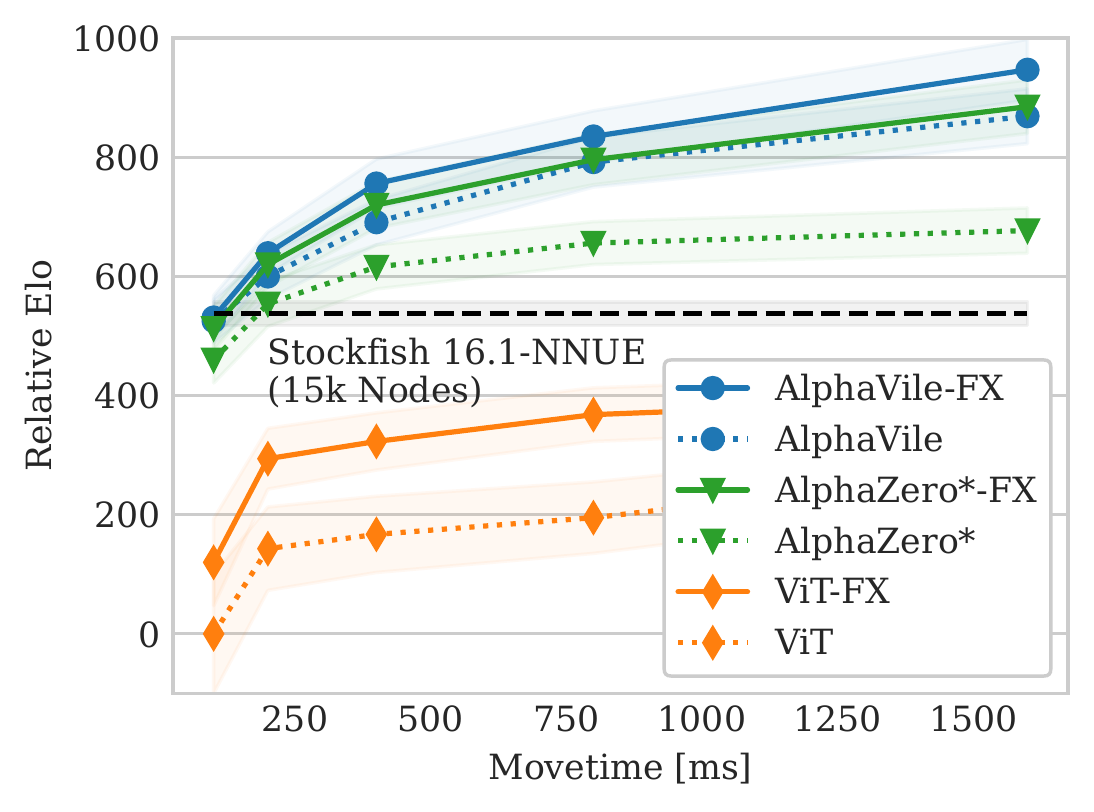}
    %\caption{Performance evaluation in chess.}
    %\caption{The \AlphaZero-FX network significantly outperforms the vanilla version that uses input representation version 1 and no WDLP head. The \AlphaVile network approaches \AlphaZero network in strength at higher move times.}
    \label{fig:strength_comparision_chess}
}
\hfill
% crazyhouse
% AlphaVile-FX, AlphaZero-FX, ViT-FX
\subfloat[Assessment of playing strength in crazyhouse.]{
    \centering
    \includegraphics[width=\figwidth\linewidth]{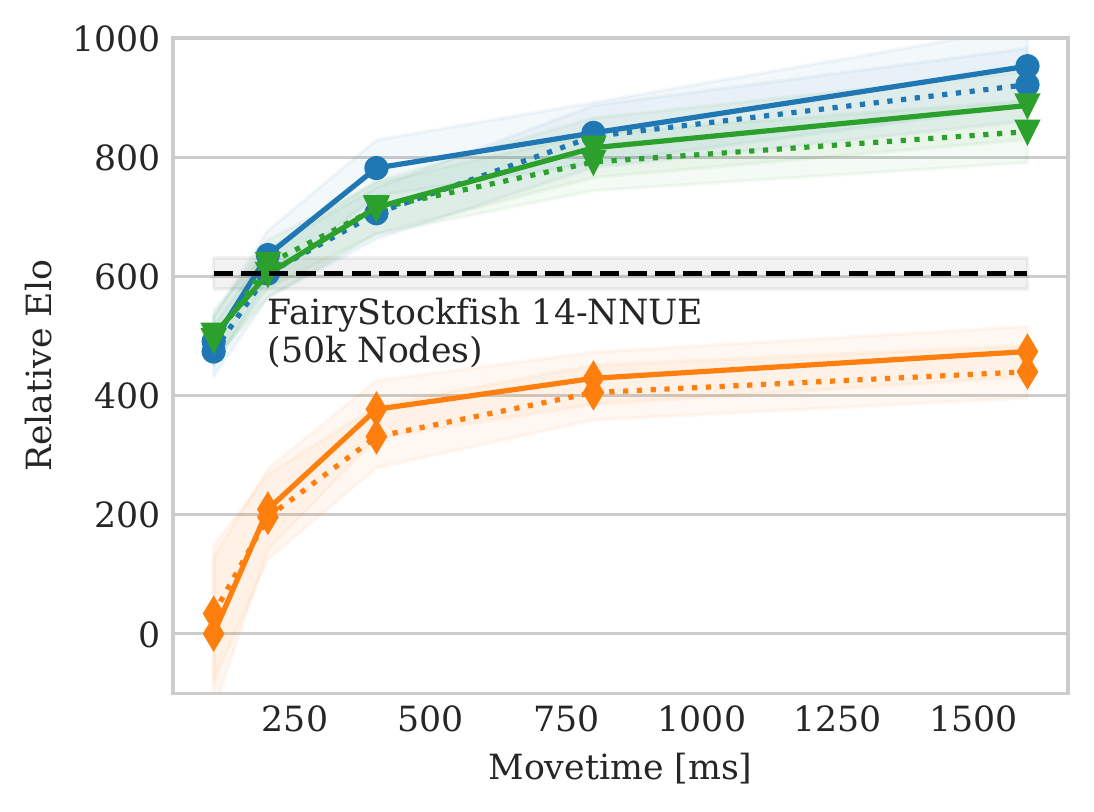}
    %\caption{Assessment of playing strength in crazyhouse.}
    %\caption{The FX networks still outperforms the vanilla versions in crazyhouse that uses input representation version 1 and no WDLP head. However, the advantage is only 30-68 Elo.}
    \label{fig:strength_comparision_crazyhouse}
}
\hfill
\subfloat[Comparative analysis of playing strength in atomic chess.]{
    \centering
    \includegraphics[width=\figwidth\linewidth]{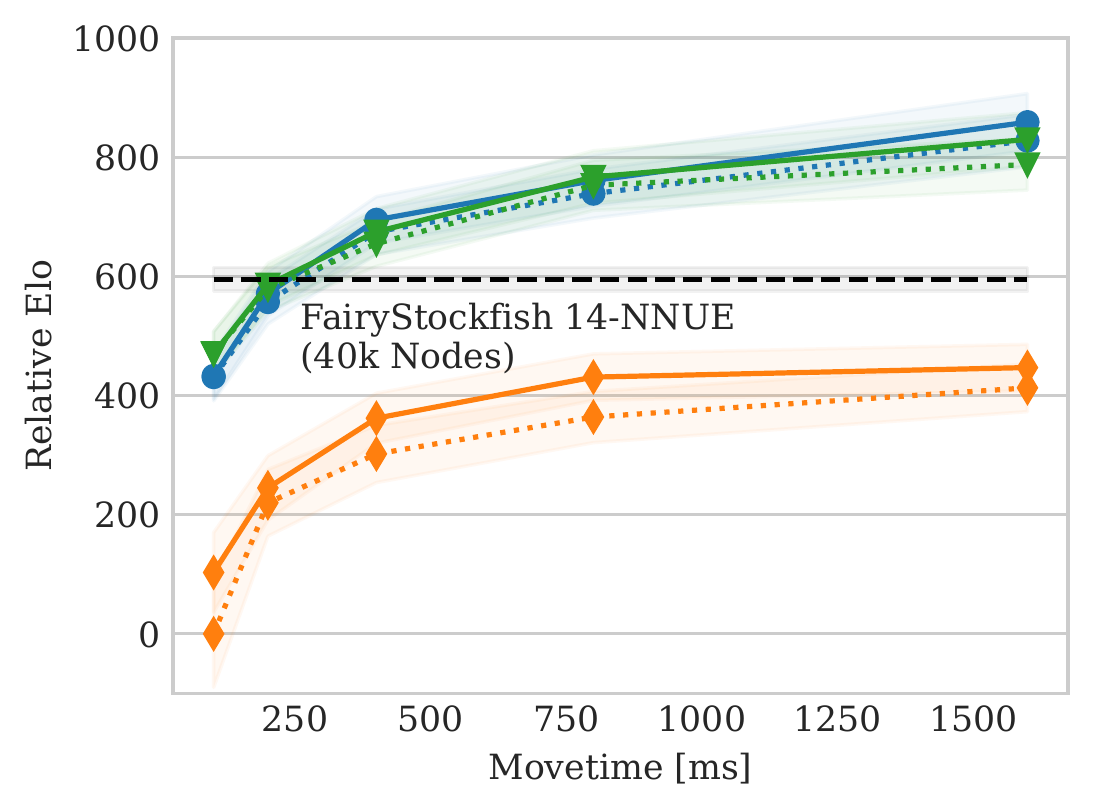}
    %\caption{Comparative analysis of playing strength in atomic chess.}
    %\caption{The FX networks still outperforms the vanilla versions in atomic that uses input representation version 1 and no WDLP head. However, the advantage is only 30-68 Elo.}
    \label{fig:strength_comparision_atomic}
}
\caption{The \AlphaZero-FX network showcases excellent performance in chess~(\ref{fig:strength_comparision_chess}), crazyhouse~(\ref{fig:strength_comparision_crazyhouse}), and atomic chess~(\ref{fig:strength_comparision_atomic}), surpassing the vanilla version using Input Representation Version 1 without the WDLP head. The performance increase in chess is noteworthy, with an increase of 180 Elo point. The performance level of the \AlphaVile network is comparable to that of the \AlphaZero network, especially at longer move times.}
\label{fig:strength_comparision}

\end{figure}

\subsection{Comparative assessment of playing strength}
\label{sec:playing_strength}
In order to evaluate the playing strength of our ViT models, we conducted a comprehensive round-robin tournament, in which we pitted AlphaVile, ViT, and AlphaZero* against each other. AlphaZero* here relates to a reimplementation of AlphaZero in the form of ClassicAra using the same model architecture as AlphaZero. The results of this tournament are graphically illustrated in Figure~\ref{fig:strength_comparision}. The ViT model %although a promising architecture in other contexts,
falls short of matching the playing strength achieved by AlphaZero* and \AlphaVile. This outcome is consistent with the results presented in Figure~\ref{fig:final_performance}, which highlights the computational disparities between ViTs and our other models.
\AlphaVile-FX slightly outperform the AlphaZero*-FX version by about 30 Elo.
%A notable finding is the emergence of the optimised AlphaZero-FX model, utilising the expanded input and loss representations.
%When dealing with restricted time per move, our AlphaZero models possess a distinct advantage over the AlphaVile and ViT models.
Elo is a metric that measures the relative playing strength difference.
We set our baseline Elo rating to 0 Elo, which refers to the weakest participant, here ViT, in our tournament.
We refrain from using a baseline Elo rating from ClassicAra from engine rating lists, because we were testing on a different hardware than used for creating the engine list.
ClassicAra-1.0.1 participated at the Top Chess Engine Championship (TCEC) season 23 and achieved an Elo rating of 3279, compared to Stockfish-dev16 (3625), LCZero-0.30 (3599).
We also add Stockfish 16.1-NNUE (15k nodes per move) and FairyStockfish 14-NNUE (50k nodes and 40k per move) as horizontal lines to Figure~\ref{fig:strength_comparision} to make the results more comparable.
The modifications to the input and loss representations in our study are substantial, leading to a significant increase in playing ability. In particular, the modification improves the performance of AlphaZero* by 180 Elo points in chess. A lessened increase is evident in chess variants, such as crazyhouse (Figure~\ref{fig:strength_comparision_crazyhouse}) and atomic chess (Figure~\ref{fig:strength_comparision_atomic}).
This emphasizes the significance of these changes, which are apparent in the enhanced playing strength across chess variants.
%We also train additional chess variants for both AlphaZero and our model to see its effects in different environments.
%For our training data, we use here the openly available lichess.org data set.
%As can be seen in Figure~\ref{fig:strength_comparision}, the updated AlphaZero model is consistently stronger than the vanilla version. The \AlphaVile model overtakes the vanilla AlphaZero model in strength and approaches the updated AlphaZero model.
%The standard ViT model, however, is considerably worse than all other competitors.
%the improvement is consitent and variies from an Elo difference from X Elo in variant Y to Z Elo in variant C.
%All games were played in controlled conditions, with each player given a fixed amount of time in milliseconds per move.
%The time limit ranged from 0.1s to 1.6s, allowing for a thorough evaluation of the models under different temporal constraints. 
Opening suites were incorporated into the gameplay to introduce a range of game scenarios. %This approach ensured a comprehensive assessment of the model's performance across a variety of game configurations.

\begin{figure*}[ht!]
    \centering
\subfloat[Analysis of average feature importance for the default input representation.]{
    \centering
     \includegraphics[width=.85\linewidth]{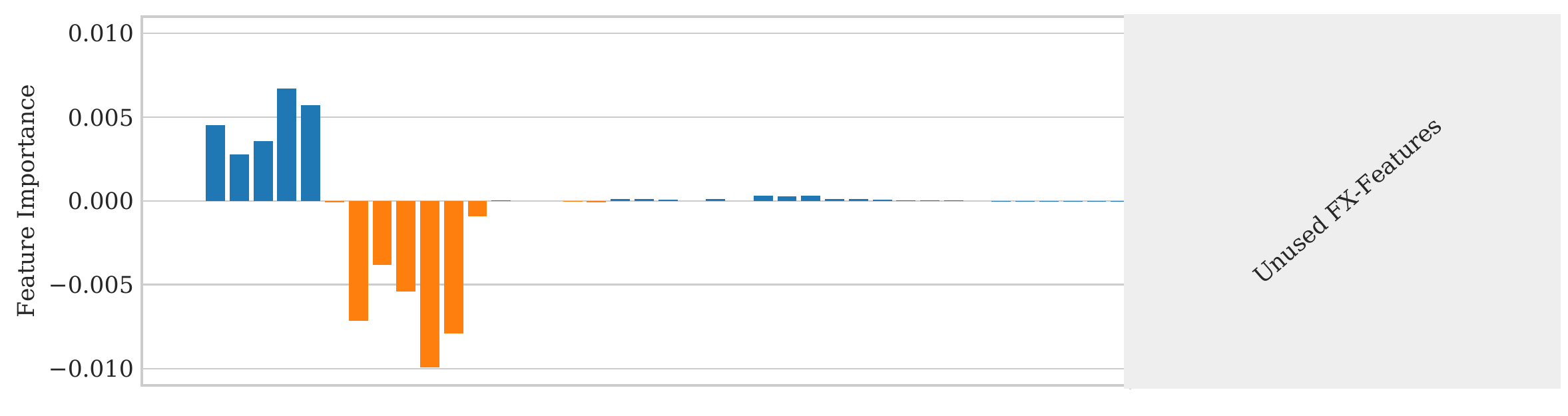}
    %\caption{Analysis of average feature importance for the default input representation.}
    \label{fig:feature_importance_v1}
}

\subfloat[Analysis of the average importance of FX features.]{
    \centering
     \includegraphics[width=.85\linewidth]{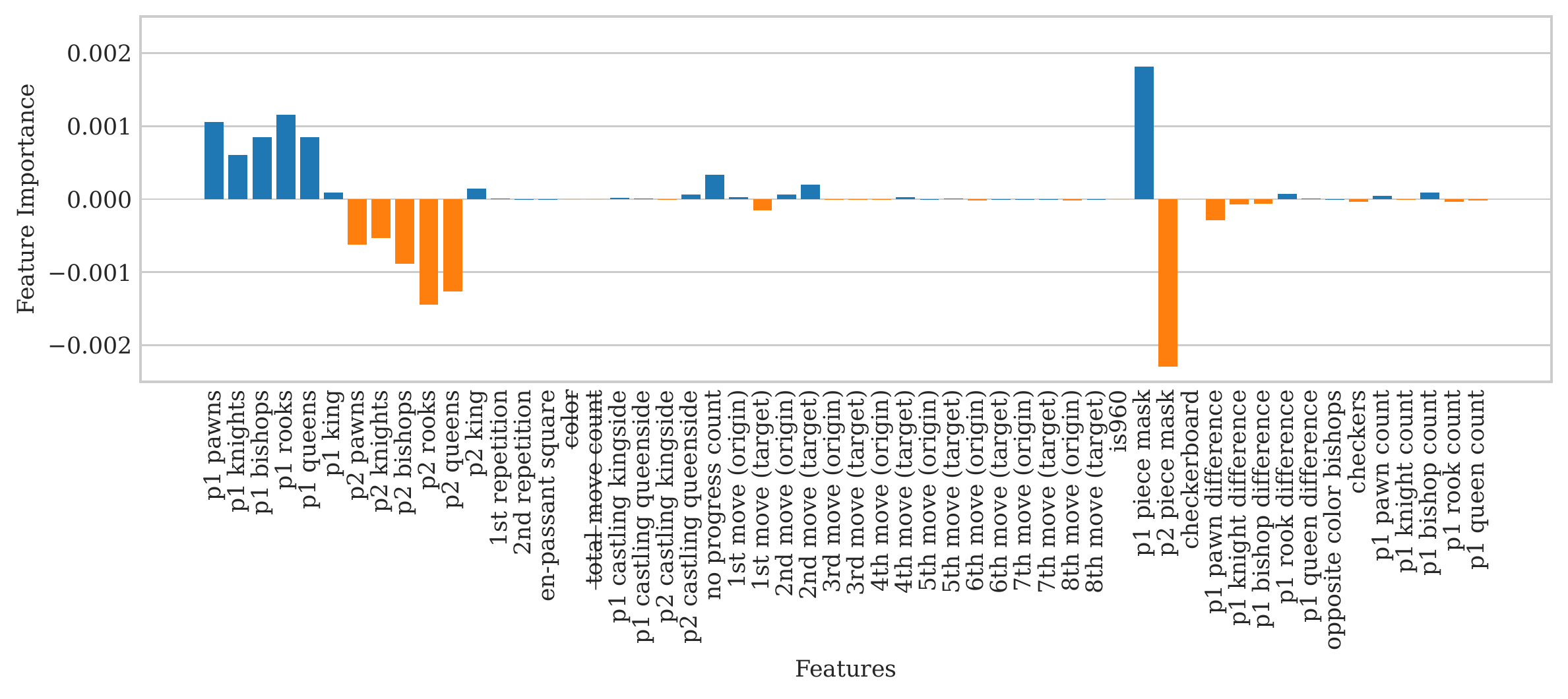}
    %\caption{Analysis of the average importance of FX features.}
    \label{fig:feature_importance_v3}
}
\caption{The newly introduced FX-features demonstrate significant usage, highlighted by the Integrated Gradients (IG) method for feature importance analysis. In the conventional input representation (\ref{fig:feature_importance}a), both positive and negative feature attributions are predominantly related to piece maps. In the enhanced input representation presented in (\ref{fig:feature_importance}b), supplementary features are incorporated, while two features marked with \sout{strike-through} are omitted. The IG method uses the average of all inputs as a baseline for the attribution calculation.}
\label{fig:feature_importance}
\end{figure*}

\subsection{Interpretability of FX-features}

Next, we investigate the interpretability of the FX features and their influence on the model. To determine the importance of each feature channel, we use the Integrated Gradients~(IG) method, a widely accepted technique in the field of neural networks' interpretability~\cite{sundararajan2017axiomatic}. 
We calculate the average attribution of each channel to the model's output value. It is crucial to establish an appropriate baseline for computing gradients. In Figure~\ref{fig:feature_importance}, we utilize the mean of all validation input features. 
%whilst for Figure~\ref{fig:opp_color_bishop_example_2}, an all-zero input is used as the baseline.

Our analysis shows that the newly introduced feature channels within the FX representation have significant utility.
%, as demonstrated by the IG method.
Some feature channels display positive attribution, while others exert a negative influence. This observation supports the logical assumption that a greater number of the opponent's pieces corresponds to a lower value loss. 
Our analysis suggests, interestingly, that the color channel and move history information are of limited importance, as shown in the second graph.
%Nonetheless, a slight correlation is evident in Figure~\ref{fig:opp_color_bishop_example_1}.
Additionally, it seems that the king's position is of low relevance. This may be misleading, as the king is always present, and a weak or strong king's position has either a positive or negative influence on the value target.
The IG method highlights the significance of the player~1 and player~2 masks as the most influential feature channels of the FX-Features.
Their integration, in conjunction with the refined value loss representation, significantly improves the evaluation of endgames with opposite-colored bishops, as further elaborated in the
\ifarxiv
appendix.
\else
appendix~\cite{czech2024representation}.
\fi

\section{Related Work}

Originally created for supervised learning tasks, transformers have been widely adopted across multiple domains, such as natural language processing and computer vision. In the realm of RL, where sequential decision-making is crucial, the usage of transformers has become an active area of research \cite{janner2021sequence}. The primary goal of combining transformers and RL is to model and, in several instances, improve decision-making by utilizing attention mechanisms. 
The sequential nature of RL tasks makes transformers a flexible framework for addressing them. Several approaches have been suggested to overcome the gap between transformers and RL, including enhancements in architecture and trajectory optimization strategies \cite{hu2023transforming}. 

Currently, the Trajectory Transformer~\cite{janner2021sequence} and Decision Transformer~\cite{chen2021decision} are among the prominent paradigms. For a thorough comprehension of the combination of transformers and RL, we suggest studying the insightful surveys by Li et al.~\cite{li2023survey} and Hu et al.~\cite{hu2023transforming}.

%This lively area highlights the potential of transformers for RL, and our work contributes to this expanding research by dealing with the particular challenges arising in the context of chess AI.
%For Final Rise, we use version 5 which has the best performance for the normal model

Recent work \cite{farebrother2024stop} has shown that utilizing classification-based approaches is generally superior to alternative regression-based approaches. 
Game-specific features have also been found to be beneficial for the game of Go~\cite{wu2019accelerating}.

The utilization of transformers in chess-related tasks has previously been explored in the literature \cite{DeLeo_2022,noever2020chess,toshniwal2022chess}, albeit these investigations differ significantly from our approach.
Previous studies mainly utilized Large Language Models (LLMs) and analyzed chess problems from a linguistic perspective. They particularly relied on techniques like Portable Game Notation (PGN) and area-specific terminology to represent chess positions textually.
A recent paper \cite{ruoss2024grandmaster} explores the utilization of transformer without search and achieves grandmaster-level performance.
Although these attempts succeeded in teaching the regulations of chess, they did not achieve the playing skills demonstrated by AlphaZero.
In another line of research, the use of transformers in chess has also been envisioned to produce annotations on chess positions, with the objective of enhancing the comprehensibility and traceability of chess analysis \cite{lee2022improving}.
%Last but not least, in the pioneering work ~\cite{steingrimsson2021chess}, the importance of feature representations was proven by performing rigorous experiments on a new advanced "System 2" type chess task, that SOTA chess architectures still struggle with. There the best performing architecture, which included just like in our experiments, both a moves-left and WDL heads, outperformed significantly larger neural networks. Furthermore, in  ~\cite{steingrimsson2021chess} the importance of behavioral diversity achieved by „exploring the agent state space and having diverse agents with a heterogenous skill set” resulting in different agents that play differently from one another was first pointed out. By giving each agent a reward for exploring different parts of the state space, a variety of chess strategies and styles which could be described as “creativity” could be achieved. Our approach where we gave our agent access to broader information was motivated by this approach.
\citet{steingrimsson2021chess} demonstrated through rigorous experiments on an advanced chess benchmark which SOTA chess architectures still struggle with, the crucial role of neural architecture improvements. This consisted of broader output variables. They also emphasized the importance of behavioral diversity among agents, which led to creative and varied chess strategies and experiments with additional heads. In this work, we take the next step, exploring input features and their representation.

\begin{comment}
In the current SOTA landscape in AI for chess, two predominant engine architectures have emerged as leading examples. The first type of architecture involves agents that expand upon the foundational principles of AlphaZero. Leela Chess Zero (Lc0) has gained prominence as an exemplar of this category.

Currently, the field of chess AI is still dominated by Stockfish.
%, a durable chess engine known for its longevity and adaptable strategies. 
Stockfish has recently undergone a transition to the NNUE (Efficiently Updatable Neural Network) architecture, which is a significant breakthrough in neural networks that was originally introduced by Yu Nasu~\cite{nasu2018efficiently} in the context of shogi, a Japanese version of chess. The NNUE architecture enables more efficient weight updates while playing.
%, leading to better shogi performance.
Although NNUE is a specialized architecture, it is primarily used by the shogi and chess communities.
%In recent years, the AlphaZero-inspired approaches and the NNUE architecture have dominated the computer chess field. 
The Top Chess Engine Championship held from 2019 to 2023 highlights the significance of these methods, demonstrating their exceptional performance and competitive advantage.
\end{comment}

% Other papers trying to use Transformer in Chess, Other papers using Transformers in RL 
\section{Conclusion}
Our study has shown that an optimized input representation and value loss definition significantly enhance the playing strength of chess AIs. Despite the prevailing belief that feature engineering has decreased in relevance with the emergence of deep learning networks, our findings challenge this assumption.
%An enhanced input representation is a crucial asset, as demonstrated by the impressive outcomes of NNUE (Efficiently Updatable Neural Network) architectures, which differ from deep residual networks by implementing shallow multi-layer perceptrons as well as the improvement with our FX representation.
%Our research confirms that this principle applies beyond deep convolutional networks to ViTs, thus demonstrating the widespread value of these enhancements.
Our new input representation includes novel characteristics that arise from the combination of existing features, including material difference and material count. Additionally, there are implicit features derived from the basic rules of chess, such as pieces giving check and the identification of bishops of opposite colors.
%Material difference characteristics have historically been fundamental components in manually crafted chess evaluation functions.
\begin{comment}
By granting the neural network direct access to these characteristics, we permit it to concentrate on more detailed and theoretical elements of the game position, rather than expending energy on creating these characteristics autonomously.
\end{comment}
%Although a simplified input representation has advantages because it can be used universally, we have found that including new features and user-defined biases has successfully accelerated the learning process. 
%Our research highlights the persistent importance of feature engineering in the field of deep learning and large transformer models, especially in the context of developing chess AI. (repetition)
%
Transformers are a versatile tool for AI recognized for their ability to process global features and effectively handle extended input sequences, thanks to their use of attention mechanisms. However, their applicability in specific domains such as timed competitive games, like chess, leads to unique challenges beyond accuracy. In such contexts, efficiency is paramount.
Efforts to improve the performance of ViTs in chess AI by fusing them with CNNs aimed to exploit the latter's efficient pattern recognition capabilities. Addressing the latency issues typically associated with transformers, these hybrid models generated slightly superior results compared to the pure convolutional network baseline, \AlphaZero.
%Acknowledging that ViTs are just one aspect of the broad range of transformer architectures is important. Options such as Decision Transformer \cite{chen2021decision} may provide distinct perspectives and results.
Furthermore, custom-made transformers that cater to the specific requirements of chess may improve performance. Ongoing experiments carried out by the Lc0 developer team in this area demonstrate potential, although further study is required and is outside the scope of this paper. We maintain that transformers hold substantial promise for advancing the field of computer chess. Particularly, their potential applications in areas such as multimodal inputs \cite{xu2022multimodal} and retrieval-based approaches \cite{borgeaud2022improving} may open new avenues for enhancing the capabilities of computer chess engines.
%Our objective is not to provide definitive conclusions, but rather to highlight the complex and intricate process of integrating transformers into chess algorithms. This is a challenging task, but it presents an appealing avenue for exploration.
%Its primary conclusion is clear: the selection of representation is crucial. Modifications made to the input features and loss function resulted in a considerably greater enhancement in chess performance than the replacement of ResNet with a transformer. 
Our findings underscore the enduring importance of feature engineering, negating any suggestion of its becoming obsolete and proposing that it remains ``forever young''.

%% file: tables/input_features_v2.tex
\begin{table*}[t]
\centering
\caption{Plane-based Feature Representation for Chess (Inputs V1.0). Features are encoded as binary maps, and specific features are indicated with $\ast$ as single values applied across the entire $8\times 8$ plane. The historical context is captured as a trajectory spanning the last eight moves. The table begins with traditional input features (listed above the horizontal line). The extended input representation (Inputs V.2.0) incorporates additional features below the horizontal line, while omitting two features marked with \sout{strike-through}.}
% Note: Version 2 is indicated in CrazyAra as Version 3.0
%\vspace{.2cm}
\label{tab:inputs_v1}
\label{tab:inputs_v2}
\begin{tabular}{lccl}
\toprule
%\begin{tabular}{l|c|c|l}
%
\label{tab:input_representation}
\textbf{Feature} & \textbf{Planes} & \textbf{Type}                         & \textbf{Comment}                                                            \\ \midrule
P1 pieces                          & 6                                & bool                      & order: \{\texttt{PAWN}, \texttt{KNIGHT}, \texttt{BISHOP}, \texttt{ROOK}, \texttt{QUEEN}, \texttt{KING}\}                                     \\
P2 pieces                          & 6                                & bool                      & order: \{\texttt{PAWN}, \texttt{KNIGHT}, \texttt{BISHOP}, \texttt{ROOK}, \texttt{QUEEN}, \texttt{KING}\}                                     \\
Repetitions\textsuperscript{*}                      & 2                                & bool                      & \multicolumn{1}{l}{how often the board positions has occurred} \\
%P1 pocket count\textsuperscript{*}                  & 5                                & \multicolumn{1}{c}{int} & order: \{\texttt{PAWN}, \texttt{KNIGHT}, \texttt{BISHOP}, \texttt{ROOK}, \texttt{QUEEN}\}                                           \\
%P2 pocket count\textsuperscript{*}                  & 5                                & \multicolumn{1}{c}{int} & order: \{\texttt{PAWN}, \texttt{KNIGHT}, \texttt{BISHOP}, \texttt{ROOK}, \texttt{QUEEN}\}                                           \\
%P1 Promoted Pawns                 & 1                                & bool                      & indicates pieces which have been promoted                                    \\
%P2 Promoted Pawns                 & 1                                & bool                      & indicates pieces which have been promoted                                    \\
En-passant square                 & 1                                & bool                      & the square where en-passant capture is possible                        \\
\sout{Color}\textsuperscript{*}                           & 1                                & bool                      & all zeros for black and all ones for white                                                   \\
\sout{Total move count}\textsuperscript{*} & 1 & int & integer value setting the move count (UCI notation)\\
%(Colour\textsuperscript{*}                           & 1                                & bool                      & all zeros for black and all ones for white)                                                   \\
%(Total move count\textsuperscript{*} & 1 & int & integer value setting the move count (uci notation))\\
%Total move count\textsuperscript{*}                 & 1                                & int                      & sets the full move count (FEN notation)                                         \\
P1 castling\textsuperscript{*}                      & 2                                & bool                      & binary plane, order: \{\texttt{KING\_SIDE}, \texttt{QUEEN\_SIDE}\}                                      \\
P2 castling\textsuperscript{*}                      & 2                                & bool                      & binary plane, order: \{\texttt{KING\_SIDE}, \texttt{QUEEN\_SIDE}\}                                      \\
No-progress count\textsuperscript{*}                & 1                                & int                      & sets the no progress counter (FEN halfmove clock)                                \\
Last Moves &	16 & bool &	origin and target squares of the last eight moves \\
is960\textsuperscript{*}  &	1 & bool  & if the 960 variant is active \\
\midrule
%\hline
%\rule[0.5ex]{1cm}{0.4pt}
%\hdashline % cause problems with sunit
{P1 pieces} &	{1} & {bool} &	grouped mask of all P1 pieces \\
{P2 pieces} &	{1} & {bool} &	grouped mask of all P2 pieces \\
{Checkerboard} &	{1} & {bool} &	chess board pattern \\
{P1 Material difference\textsuperscript{*}} &	{5} & {int} &	{order: \{\texttt{PAWN, KNIGHT, BISHOP, ROOK, QUEEN}\}} \\
%, normalized by 8, + means positive, - means negative \\
{Opposite color bishops\textsuperscript{*}} & {1} & {bool} &	{if they are only two bishops of opposite color} \\
{Checkers} &	{1} & {bool} &	{all pieces giving check} \\
{P1 material count\textsuperscript{*}} &	{5} & {int} &	{order: \{\texttt{PAWN, KNIGHT, BISHOP, ROOK, QUEEN}\}} \\ %, normalized by 8 \\
\midrule
Total                             & 39 / 52                               &                              &                                                                                           \\
\bottomrule
\end{tabular}
\end{table*}

%% file: supplementary.tex
\clearpage
\section{Supplementary Materials}

\subsection{Final Performance Overview of AlphaVile}

\begin{table*}[t]
\centering
\caption{A Comparison of Neural Network Architectures Using Extended Input Representation and WDLP Value Loss Formulation. The largest AlphaVile model provides the most favorable results, but with higher latency. The normal-sized AlphaVile model achieves comparable accuracy and latency to the ResNet architecture.}
\label{tab:final_performance}
\begin{tabular}{lcccccc}
\toprule
\textbf{Network Architecture} & \textbf{Combined Loss} & \textbf{Policy Acc. (\%)} & \textbf{Latency ($\mu s$)} & \textbf{Flops} \\
\midrule
\AlphaZero-FX \cite{silver_mastering_2017}       & 1.1673  $\pm$ 0.0040      & 59.43 $\pm$ 0.09  & 68.25 &  1.494\,G    \\
%ResNet-SE (AlphaZero)    & 17    & TODO      & 1.16195 $\pm$ 0.00014     & 59.633 $\pm$ 0.047  & 11,729       \\
%RISE-v2 \cite{czech2020learning}    & 16    & TODO      & 1.20274 $\pm$ 0.0016      & 58.367 $\pm$  0.047  & 25,326      \\
LeViT-FX \cite{graham2021levit}      & 1.2596 $\pm$ 0.0040       & 56.93 $\pm$ 0.09  & 57.90 &  0.413\,G     \\
NextViT-FX \cite{li2022next}      & 1.1725 $\pm$ 0.0040      & 59.10   $\pm$ 0.08  & 54.59  &  0.364\,G  \\
ViT-FX \cite{dosovitskiy2020image}      & 1.6866 $\pm$ 0.0014      & 47.40  $\pm$ 0.16  & 70.72   &   20.82\,M \\%20.32M \\
{\AlphaVile-FX (large)}                  &   {1.1323 $\pm$ 0.0053}  &   {60.20 $\pm$ 0.16} &{87.15}  & 0.508\,G     \\
\AlphaVile-FX (normal)                 &   1.1531 $\pm$ 0.0037  &    59.63 $\pm$ 0.13  & 67.18  & 0.374\,G    \\
\AlphaVile-FX (small)                  &   1.1861 $\pm$ 0.0082  &    58.80 $\pm$ 0.22  & 49.66   & 0.232\,G   \\
\AlphaVile-FX (tiny)                   &   1.2193 $\pm$ 0.0101  &    57.87 $\pm$ 0.25  & 38.92  & 0.171\,G    \\
\bottomrule
\end{tabular}
\end{table*}

Table~\ref{tab:final_performance} provides a detailed report on the performance metrics of the \AlphaVile architecture compared to other efficient neural network architectures.

\subsection{Preliminary Experiments for Building AlphaVile}
The AlphaVile architecture was developed in stages, commencing with preliminary experiments to evaluate the effectiveness of utilizing pure transformer-based neural networks for standard chess. Initial results, outlined in Table~\ref{fig:final_performance}, indicated that these networks exhibited sluggish performance and relatively high rates of loss. As part of our study, we endeavored to create a hybrid architecture that integrates both convolutional and transformer components to boost both efficiency and performance.

The process involved several stages. First, we commenced by determining the appropriate convolutional base block. Afterwards, we investigated the best scaling factors for the network's depth and width. Finally, the convolutional stem and transformer block were cleverly combined to create the hybrid CNN-transformer architecture that powers AlphaVile.

In the study by Li et al.~\cite{li2022next}, various convolution blocks, such as the ``Next Convolution Block'', ConvNext, Transformer, PoolFormer and Uniformer block, are comprehensively compared. However, this comparison omitted two critical components: the classical residual block that utilizes two 3$\times$3 convolutions, observed in the conventional AlphaZero network, and the mobile convolution block that uses group depthwise convolution. Thus, our primary aim is to establish which of these blocks, portrayed in Figure~\ref{fig:conv_blocks}, functions optimally under equal latency constraints. The objective is to select this block as our standard convolutional block after conducting an initial experiment.

The data presented in Table~\ref{tab:conv_blocks} shows that the mobile convolutional block~\cite{sandler2018mobilenetv2} outperforms the classical residual block~\cite{he2016deep} to a slight extent. However, the subsequent convolutional block~\cite{li2022next} demonstrates noticeably lower performance under equivalent latency constraints.

\begin{table*}[]
\centering
\caption{Results from a grid search experiment analyzing various scaling parameters to adhere to the principles of efficient net design~\cite{tan2019efficientnet}. The experiment systematically varied the depth scaling factor $\alpha$ and channel size scaling factor $\beta$. The best performance was attained with a depth scaling factor of $\alpha$=1.8 and a channel size scaling factor of $\beta=\sqrt{\nicefrac{10}{9}}$. It is noteworthy that scaling along $\beta$ yielded a more favorable latency-to-FLOPS ratio than anticipated.}
\label{tab:scaling_orig}
\begin{tabular}{ccccccc}
\toprule
$\mathbf{\alpha}$ & $\mathbf{\beta}$ & \textbf{Depth} & \textbf{Channels} & \textbf{Combined Loss} & \textbf{Policy Accuracy (\%)} & \textbf{Latency ($\mu s$)} \\
\midrule
1.0   & $\sqrt{2}$     & 10    & 272      & 1.21497          & 58.1  & 45.40   \\%25,501 /  / 21,803  (282 channels) / 20,328 (296 channels) 
1.2   & $\sqrt{\nicefrac{5}{3}}$ & 12    & 248      & 1.21381          & 58.0  & \textbf{44.55}      \\%27,069 / / 19,508 (264 channels)
1.4   & $\sqrt{\nicefrac{10}{7}}$ & 14    & 229      & 1.19888          & 58.5 & 47.88         \\%24,989 / 
1.6   & $\sqrt{\nicefrac{5}{4}}$ & 16    & 215      & 1.20036          & 58.5  & 50.64        \\%22,559 / 
1.8   & $\sqrt{\nicefrac{10}{9}}$ & 18    & 202      & \textbf{1.19084}          & \textbf{58.6}  &  51.15 \\%\textbf{22,894} /
2.0   & 1           & 20    & 192      & 1.19559          & 58.4   & 50.20       \\ %23,105 / 
\bottomrule
\end{tabular}
\end{table*}

\begin{table*}[]
\centering
\caption{Results of a grid search experiment with scaling parameters adapted according to the principles of efficient net design~\cite{tan2019efficientnet}. The best performance was achieved with a depth scaling factor of $\alpha=1.8$ and a channel size scaling factor of $\beta={(\nicefrac{10}{9})}^{\nicefrac{5}{8}}$. It is worth noting that latency measurements remain consistent across all configurations.}
\label{tab:scaling_updated}
\begin{tabular}{ccccccc}
\toprule
$\mathbf{\alpha}$ & $\mathbf{\beta}$ & \textbf{Depth} & \textbf{Channels} & \textbf{Combined Loss} & \textbf{Policy Accuracy (\%)} & \textbf{Latency ($\mu s$)} \\
\midrule
1.0   & ${2}^{\nicefrac{5}{8}}$     & 10    & 296      & 1.21062 $\pm$ 0.00117    & 58.133 $\pm$ 0.047  & 49.19   \\
1.2   & ${(\nicefrac{5}{3})}^{\nicefrac{5}{8}}$ & 12    & 264      & 1.19977 $\pm$ 0.00263    & 58.333 $\pm$ 0.124  & 51.26      \\
1.4   & ${(\nicefrac{10}{7})}^{\nicefrac{5}{8}}$ & 14    & 240      & 1.19820 $\pm$ 0.00791          & 58.500 $\pm$ 0.216 & \textbf{49.01}        \\
1.6   & ${(\nicefrac{5}{4})}^{\nicefrac{5}{8}}$ & 16    & 221      & 1.20031 $\pm$ 0.00924          & 58.400 $\pm$ 0.216  & 51.67        \\
1.8   & ${(\nicefrac{10}{9})}^{\nicefrac{5}{8}}$ & 18    & 205     & \textbf{1.19009} $\pm$ \textbf{0.00488}         & \textbf{58.667} $\pm$ \textbf{0.169}  & 53.54 \\
2.0   & 1           & 20    & 192      & 1.19039 $\pm$ 0.00740          & 58.600 $\pm$ 0.308   & 50.20       \\
\bottomrule
\end{tabular}
\end{table*}

\begin{table*}[]
\centering
\caption{Results of a grid search exploring different ratios of $5\times5$ kernel filters. The best outcome is achieved by combining 50\,\% $5\times5$ convolutions with 50\,\% $3\times3$ convolutions, and using an adjusted expansion ratio of 2.04 instead of the traditional 3.0.}
\begin{tabular}{cccccc}
\toprule
\textbf{$\mathbf{5\times5}$ Kernel Filter Ratio (\%)} & \textbf{Depth} & \textbf{Channels} & \textbf{Combined Loss} & \textbf{Policy Accuracy (\%)} & \textbf{Latency ($\mu s$)} \\
\midrule
0    & 16    & 221      & 1.20031 $\pm$ 0.00924          & 58.400 $\pm$ 0.216  & 51.67        \\  %  1.18865
25   & 16    & 221      & 1.18016 $\pm$ 0.00594          & 59.000 $\pm$ 0.216  & \textbf{50.66}       \\
50   & 16    & 221      & \textbf{1.17336} $\pm$ \textbf{0.00472}          & \textbf{59.199} $\pm$ \textbf{0.141}  & 51.57       \\
75   & 16    & 221      & 1.18521 $\pm$ 0.00758          & 58.766 $\pm$ 0.205  & 52.41       \\
100  & 16    & 221      & 1.18435 $\pm$ 0.00514          & 58.833 $\pm$ 0.124  & 52.32       \\
\bottomrule
\end{tabular}
\end{table*}

%* Experiment B:
\subsection{Optimizing Scaling Ratios for Network}
After identifying the most suitable convolutional base block, we explore the optimal scaling ratios for our network. Our approach is in alignment with the compound scaling methodology described in the EfficientNet framework \cite{tan2019efficientnet}.

The compound scaling method, introduced in the EfficientNet framework, utilizes a compound coefficient, denoted as $\phi$, to adjust the network width, depth, and resolution in a systematic manner based on the following principles: 
\begin{multline}
\quad\quad\quad\quad\quad\quad\quad\quad \text{depth:}\;d = \alpha^\phi\\
\text{width:}\;w = \beta^\phi\\
\text{resolution:}\;r = \gamma^\phi\\
\text{s.t.} \quad \alpha \cdot \beta^2 \cdot \gamma^2 \approx 2\\
\alpha \geq 1, \beta \geq 1, \gamma \geq 1\\
\end{multline}

In our experiment, we maintain a constant input resolution of an 8$\times$8 grid, thus we set $\gamma$ to 1. However, we have noticed that increasing the width ($\beta$) does not entirely meet our criteria:
\begin{equation}
    \alpha \cdot \beta^2 \approx 2
\end{equation}
This is evident in the ``Latency ($\mu s$)'' column of Table \ref{tab:scaling_orig}, where we observe increased latency as we raise $\beta$ in comparison to $\alpha$.
%is more efficient than that 
Our results are consistent with previous studies, including Zagoruyko and colleagues' work \cite{zagoruyko2016wide}, which suggests that widening the network is more effective on GPU architectures than on CPU and is not aligned with the number of floating-point operations.

% Network
To achieve an optimal balance in response to network latency variations, we have formulated an adjusted criterion. This refined criterion is defined as follows:
\begin{equation}
    \alpha \cdot \beta^{1.6} \approx 2
\end{equation}

This adjustment aligns more accurately with our latency considerations, as shown in Table~\ref{tab:scaling_updated}. Our exhaustive grid search indicates that the optimal configuration involves an expansion ratio of $\alpha = 1.8$ and $\beta = {(\nicefrac{10}{9})}^{\nicefrac{5}{8}}$.

It should be noted that alterations to the network structure have a direct effect on latency, which can result in either an increase or decrease in latency. Therefore, this refined scaling ratio has been adopted for subsequent experiments to maintain consistent latency levels across different configurations.

%Investigating the Effect of Transformer Blocks
%In the following experiment, detailed in Table~\ref{tab:transformer_blocks}, we explore the integration of transformer blocks as proposed by~\cite{li2022next} into the convolutional architecture. We implemented a hybrid architecture following the recommendations in~\cite{li2022next}, which involves placing one transformer block after a designated number of convolutional blocks. Our research shows that integrating two transformer blocks per fifteen layers produces the most optimal performance.

\subsection{GPU Utilization}

\begin{figure}
\begin{minipage}{0.49\textwidth}
    \centering
    \includegraphics[width=0.8\linewidth]{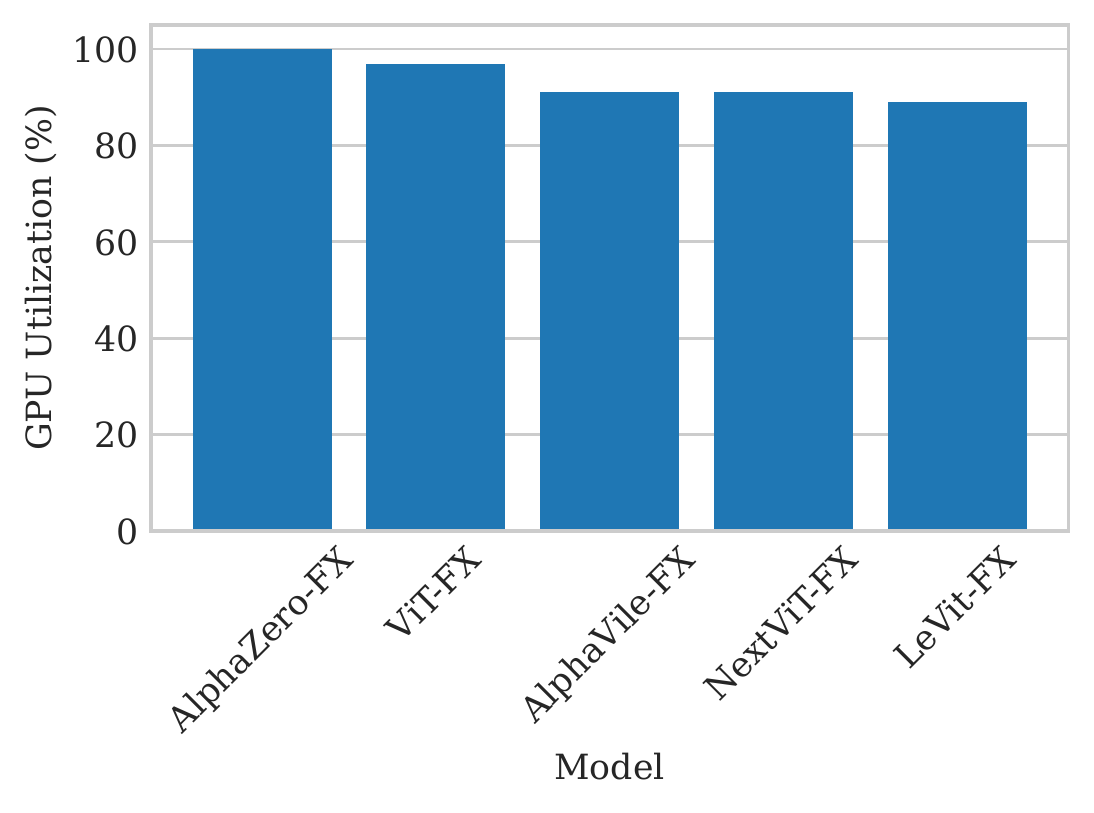}
    \caption{An in-depth analysis of GPU usage across different model structures, highlighting the exceptional performance of the standard \AlphaZero-FX network compared to its competitors.}
    \label{fig:gpu_util}
\end{minipage}
\end{figure}
    
As high GPU utilization is generally desired, it is noteworthy that the pure convolutional ResNet architecture employed by \AlphaZero shows the maximum GPU utilization, peaking at 100\,\%. On the other hand, a range of ViT architectures demonstrate GPU utilization rates between 89\,\% and 97\,\%, as shown in Figure~\ref{fig:gpu_util}.

% TODO add commentary here
\input{tables/opposite_color_bishops}

\ifarxiv
\newcommand{\chessboardscale}{0.8}
\newcommand{\featureboardscale}{0.25}
\newcommand{\wideframescale}{0.8}
\else
\newcommand{\chessboardscale}{0.9}
\newcommand{\featureboardscale}{0.3}
\newcommand{\wideframescale}{0.85}
\fi

% Show one winning and one drawing position
\begin{figure*}[htp!]
\centering
\subfloat[N. Miller vs. A. Saidy. White to move.]{
    \scalebox{\chessboardscale}{
    \chessboard[setfen=8/3k4/8/2pK4/8/4b1p1/8/5B2 w - - 0 56, showmover=false]
    }
    \label{fig:miller_saidy}
}
\hfill
\subfloat[Spatial average feature importance for the default input representation.]{
    \includegraphics[width=\featureboardscale\linewidth]{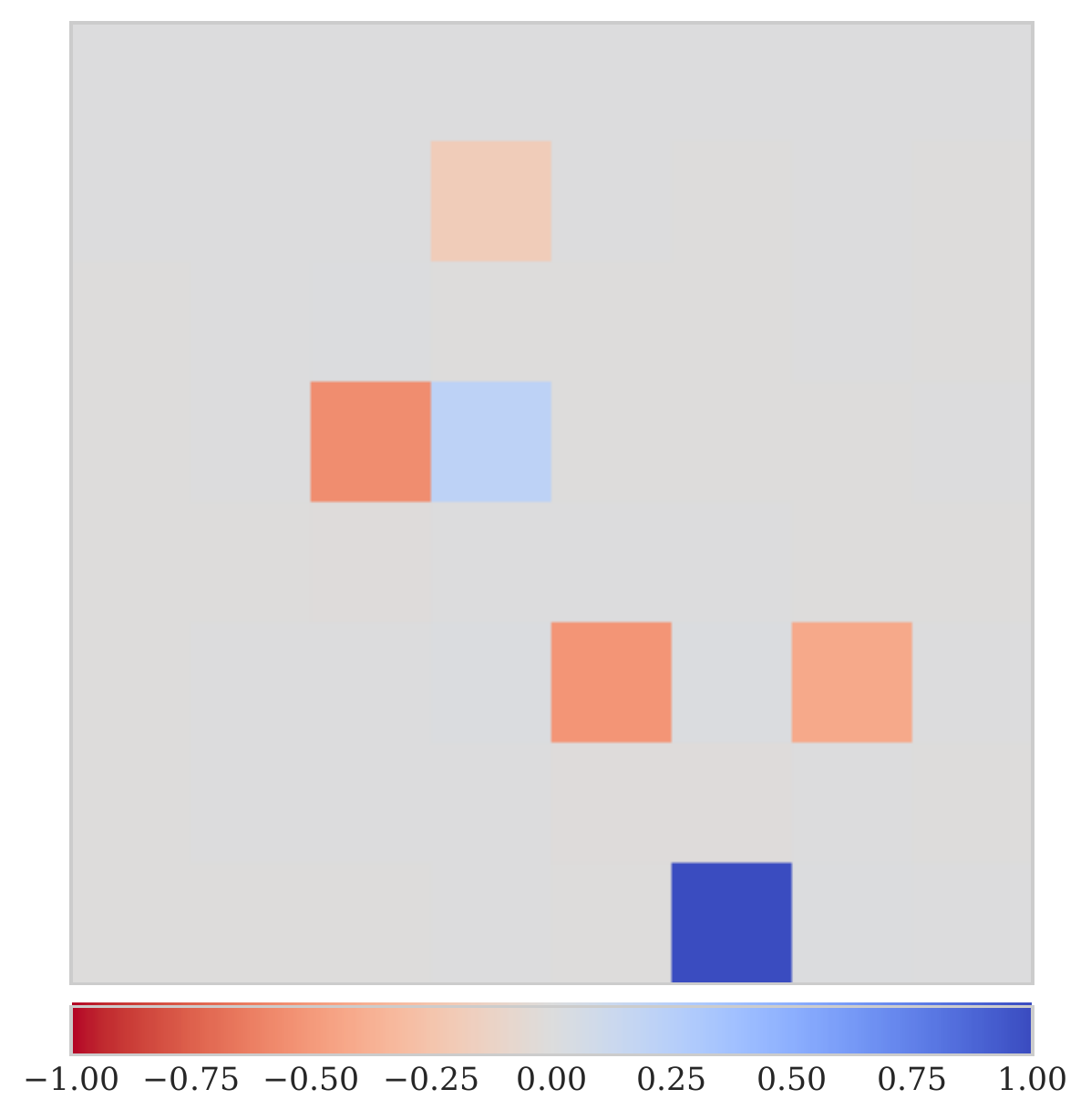}
    \label{fig:feature_importance_v1}
}
\hfill
\subfloat[Spatial average feature importance for the FX representation.]{
    \includegraphics[width=\featureboardscale\linewidth]{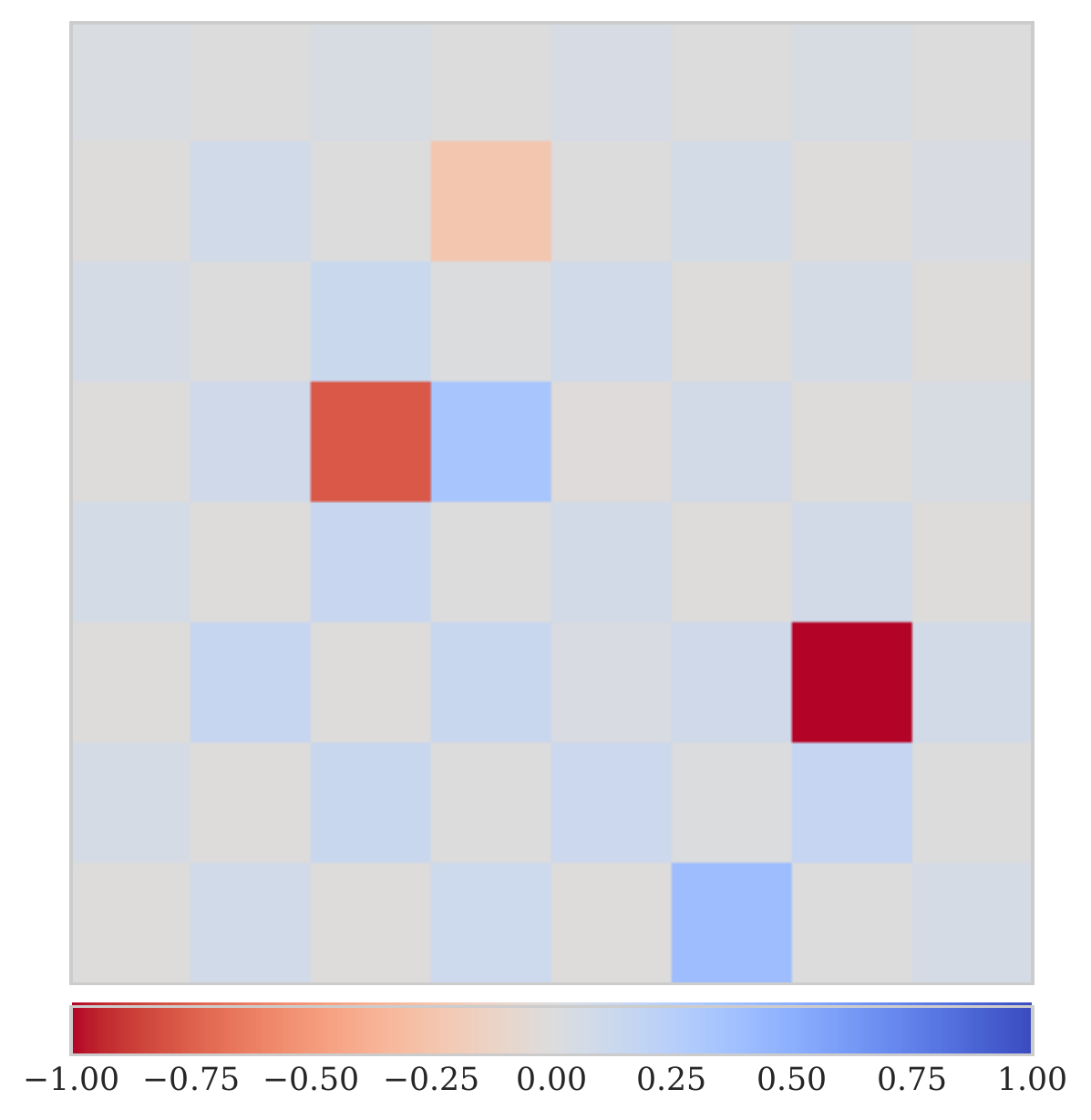}
    \label{fig:feature_importance_v3}
}
\vfill
\subfloat[Average feature importance for the default input representation.]{
    \includegraphics[width=\wideframescale\linewidth]{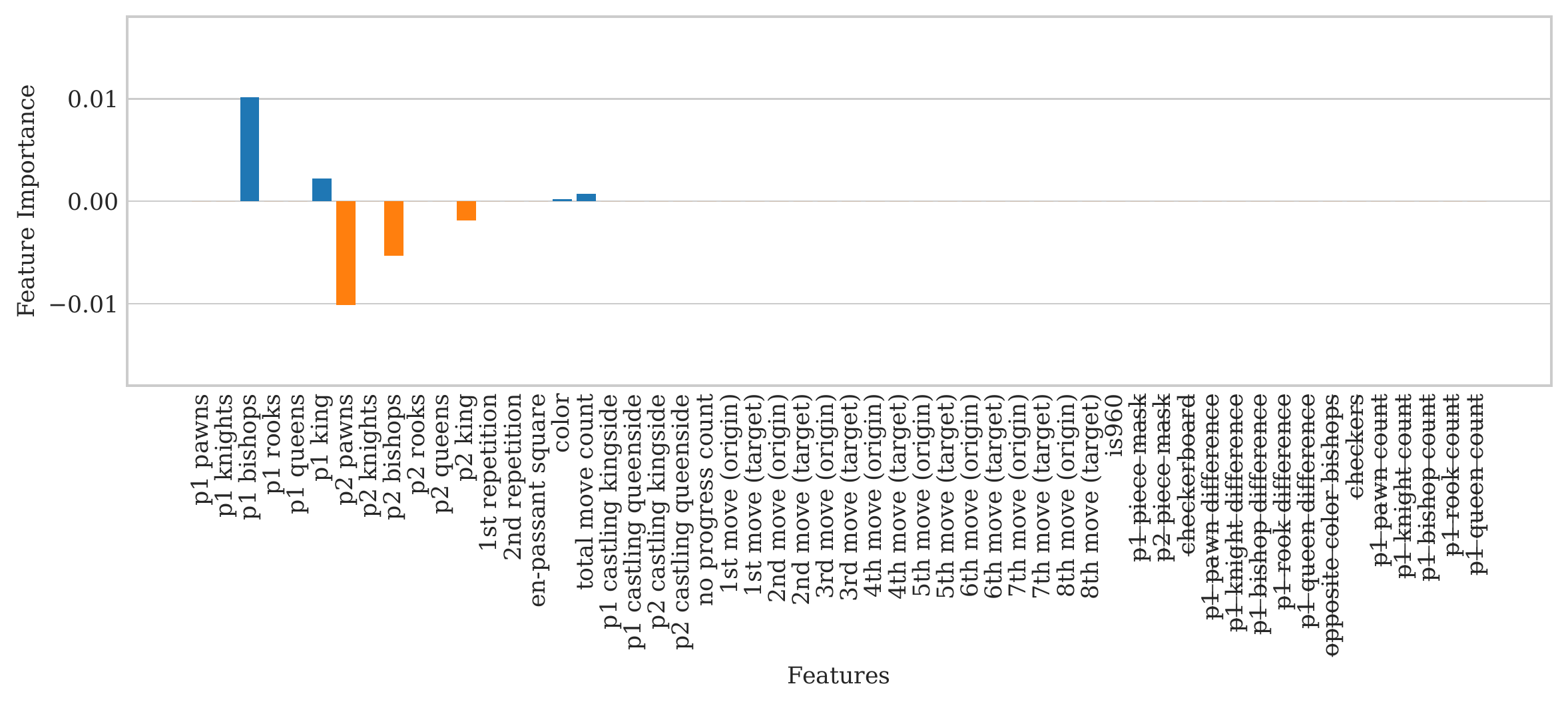}
    \label{fig:feature_importance_v1}
}
\hfill
\subfloat[Average feature importance for the FX-features.]{
    \includegraphics[width=\wideframescale\linewidth]{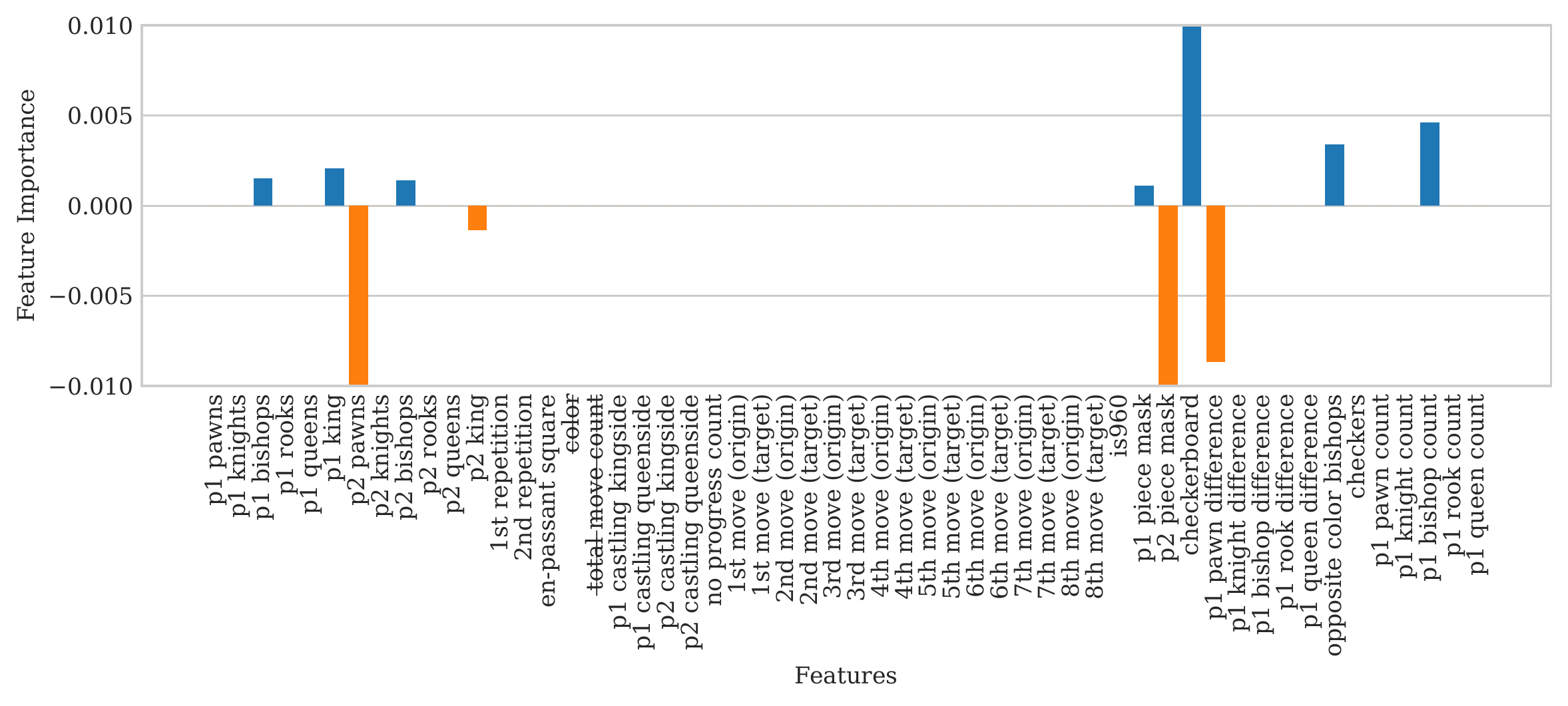}
    \label{fig:feature_importance_v3}
}
\caption{Chess endgame position from the historic game N. Miller vs. A. Saidy, 1971, along with the corresponding feature importance analyses for both the standard AlphaZero-Resnet and the AlphaZero-FX Resnet. White resigned in this drawn position (Forsyth–Edwards Notation (FEN)): \texttt{8/3k4/8/2pK4/8/4b1p1/8/5B2 w}). Single value evaluation by the default network: -0.4940, while the FX-Network evaluates it as -0.2304. Values are presented with respect to the current player to move. Integrated gradient analysis employed all-zero inputs as the baseline.}
\label{fig:opp_color_bishop_example_1}
\end{figure*}

\begin{figure*}[htp!]
\centering
\subfloat[Kotov vs. Botvinnik. Black to move.]{
    \scalebox{\chessboardscale}{
    \chessboard[setfen=8/8/4b1p1/2Bp3p/5P1P/1pK1Pk2/8/8 b - - 0 56, showmover=false]
    }
    \label{fig:kotov_botvinik}
}
\hfill
\subfloat[Spatial average feature importance for default input representation.]{
    \includegraphics[width=\featureboardscale\linewidth]{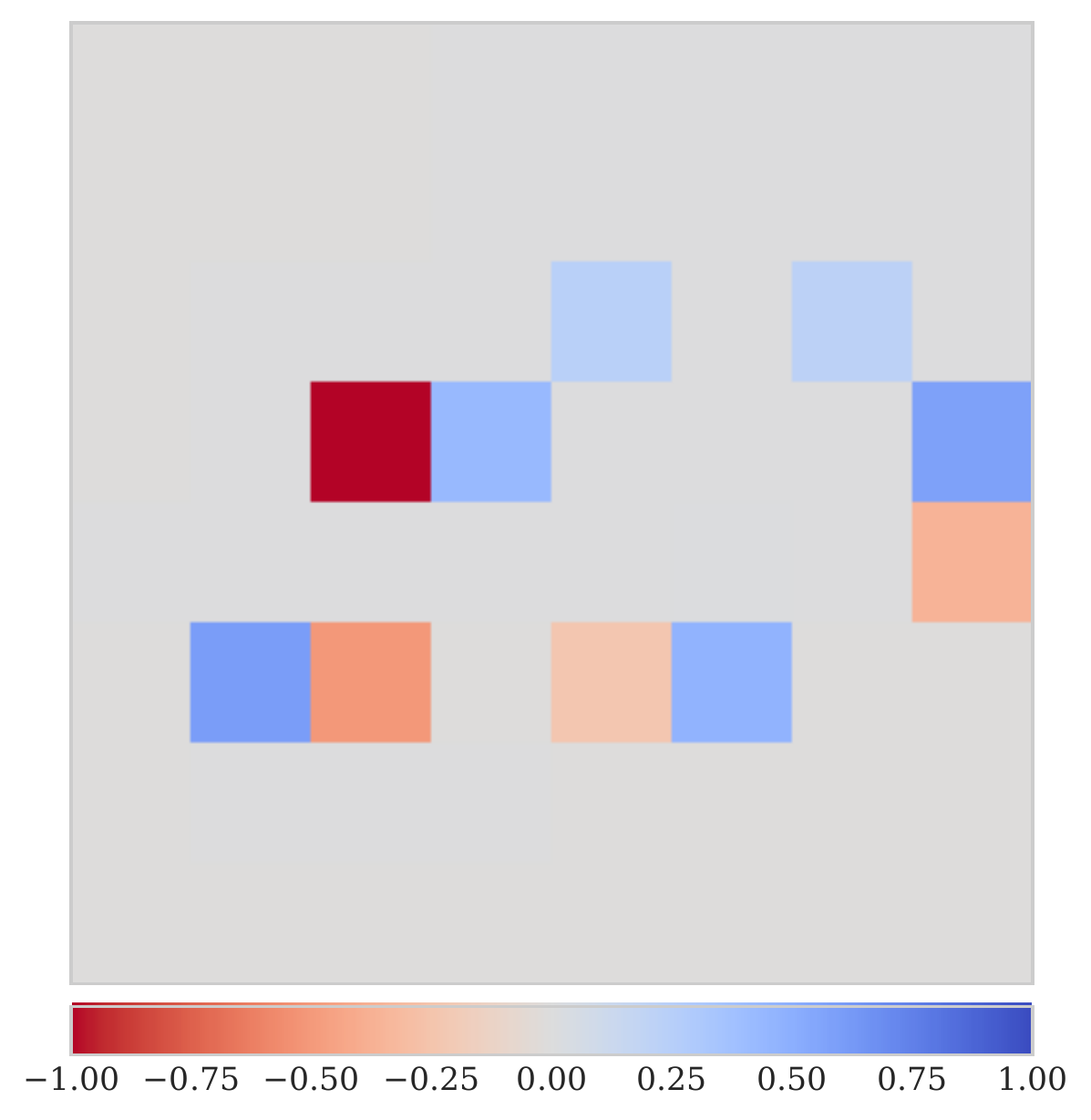}
    \label{fig:feature_importance_v1_kotov}
}
\hfill
\subfloat[Spatial average feature importance for FX-representation.]{
    \includegraphics[width=\featureboardscale\linewidth]{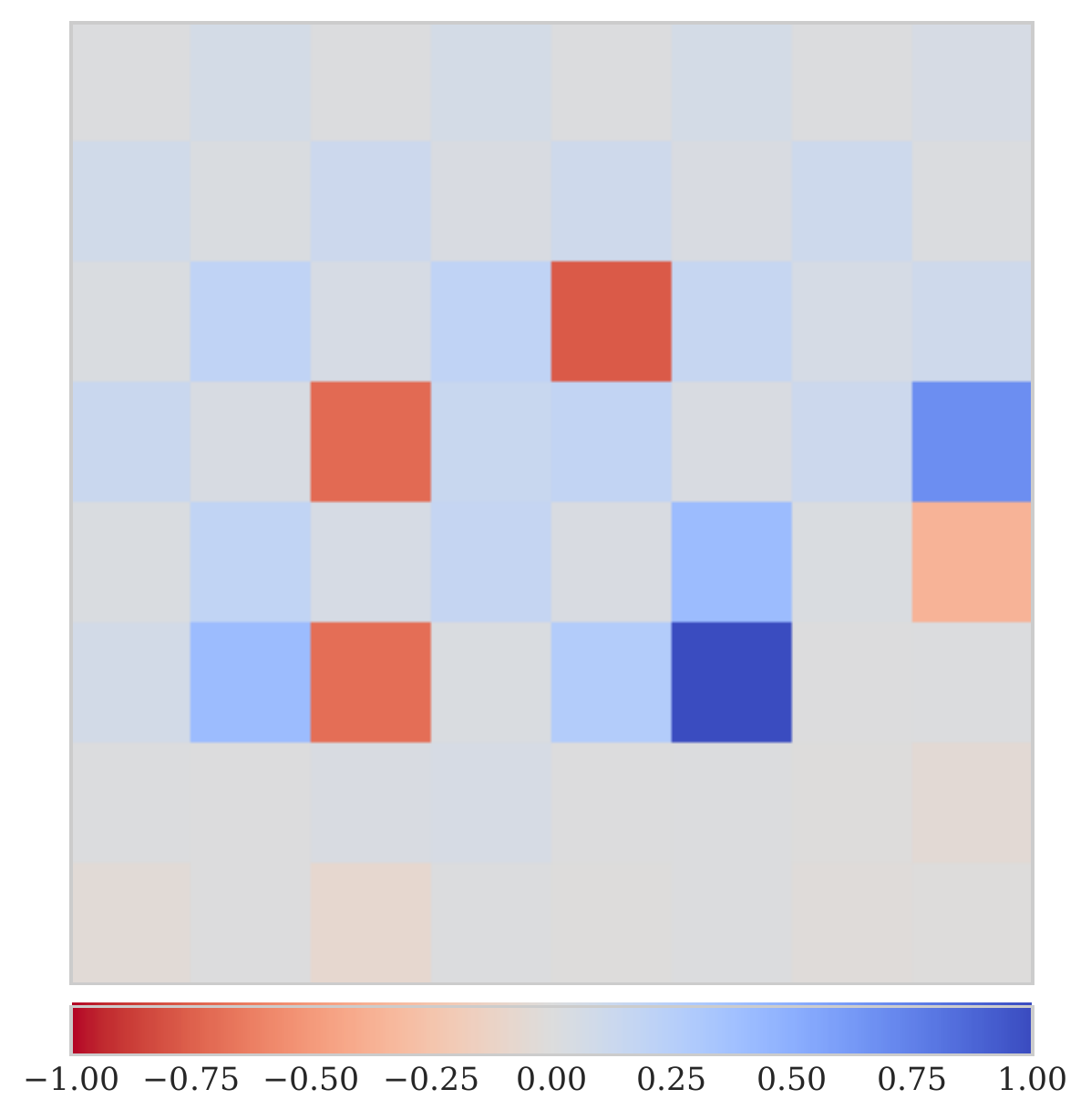}
    \label{fig:feature_importance_v3_kotov}
}
\vfill
\subfloat[Average feature importance for default input representation.]{
    \includegraphics[width=\wideframescale\linewidth]{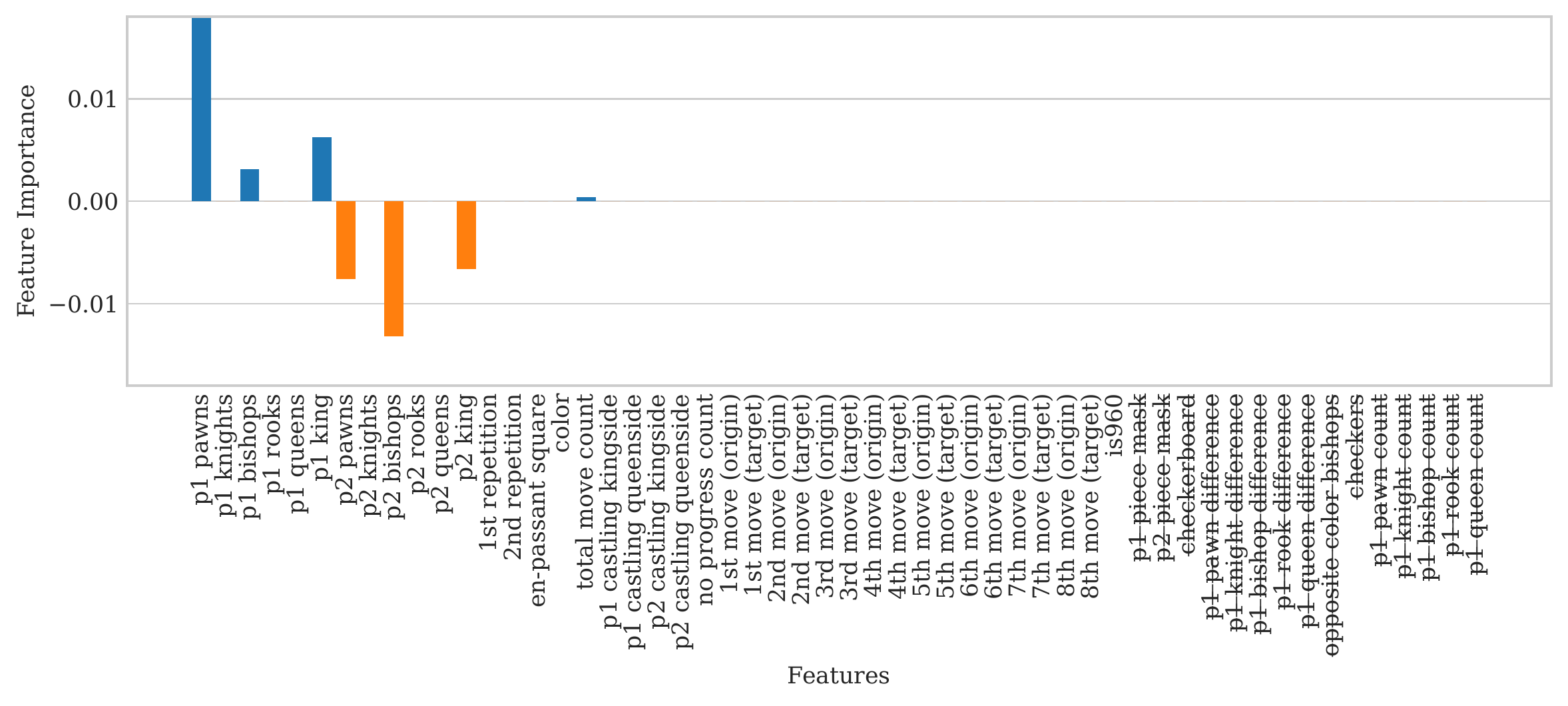}
    \label{fig:feature_importance_v1_kotov_full}
}
\hfill
\subfloat[Average feature importance for FX-features.]{
    \includegraphics[width=\wideframescale\linewidth]{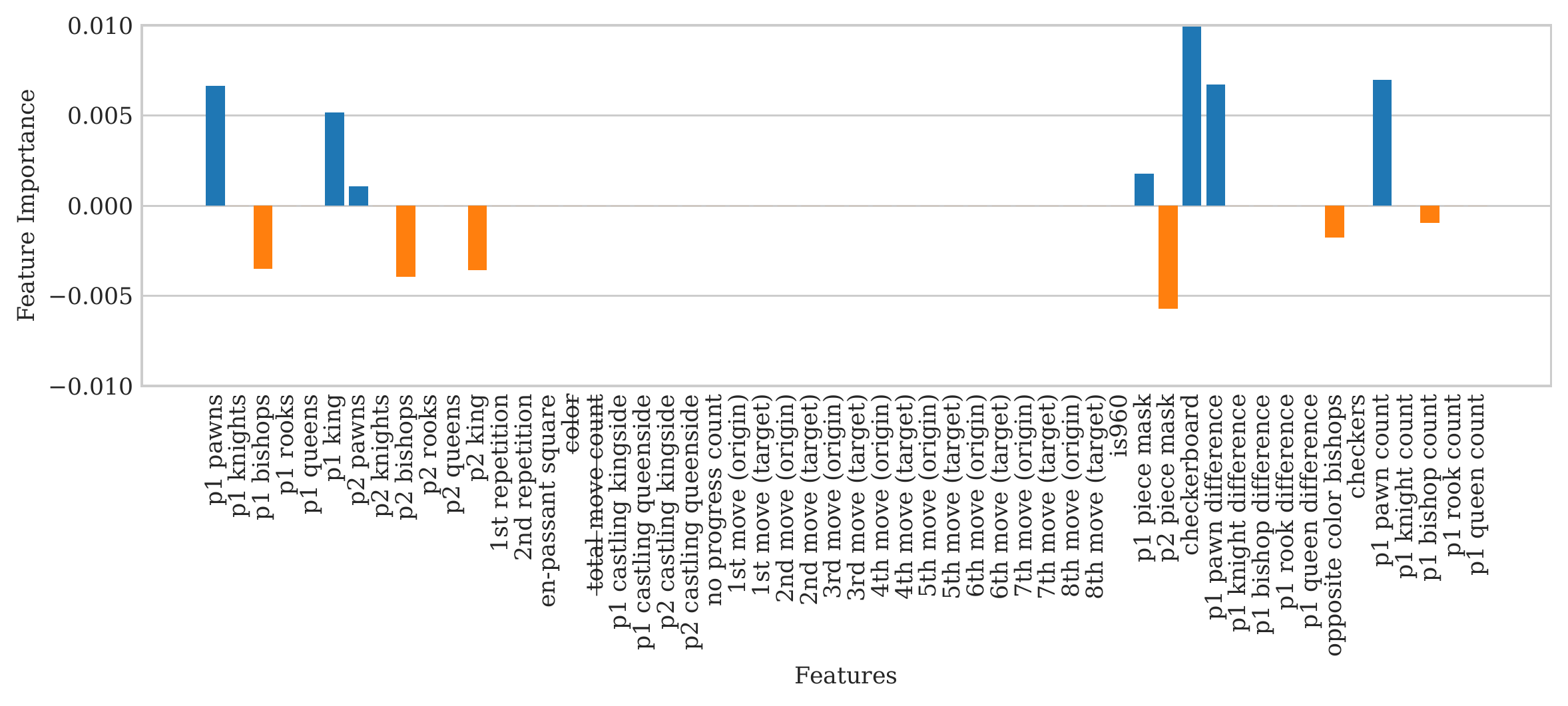}
    \label{fig:feature_importance_v3_kotov_full}
}
\caption{Visualization of a chess endgame position from the historic Kotov vs. Botvinnik game in 1955. The associated feature importance of both the standard AlphaZero-Resnet and the AlphaZero-FX Resnet models is displayed. Black secures victory by sacrificing both the g- and d-pawns, creating a new passed-pawn in the process. Forsyth–Edwards Notation (FEN): \texttt{8/8/4b1p1/2Bp3p/5P1P/1pK1Pk2/8/8 b}. Single-value evaluations by the default network: 0.4053 vs. FX-Network: 0.4225. All-zero inputs serve as the baseline for the Integrated Gradient method.}
\label{fig:opp_color_bishop_example_2}
\end{figure*}

\begin{table}[t]
\caption{Hyperparameter Configuration for Experimental Settings. This table provides a comprehensive overview of the essential hyperparameters utilised in our experimental design. }
\label{tab:hyperparams}
\centering
\begin{tabular}{llll}
\toprule
\textbf{Hyperparameter} & \textbf{Value} & \textbf{Hyperparameter} & \textbf{Value}  \\
\midrule
max learning rate       & 0.07           & value loss factor       & 0.01           \\
min learning rate       & 0.00001        & policy loss factor      & 0.988          \\
batch size              & 1024           & wdl loss factor         & 0.01           \\
max momentum            & 0.95           & plys to end loss factor & 0.002          \\
min momentum            & 0.8            & stochastic depth probability & 0.05      \\
epochs                  & 7              & pytorch version         &  1.12.0    \\
optimizer               & NAG            &&\\

\bottomrule
\end{tabular}
\end{table}

\newpage
\subsection{Case Study: Opposite Color Bishop Positions}

Opposite color bishop positions are widely recognized as complex endgame scenarios in chess, often leading to draws even when one player has a material advantage. This section aims to determine whether the use of FX representation improves the evaluation of these challenging positions.
To conduct this analysis, fully trained neural networks were subjected to a series of tests covering 20 different opposing color bishop positions. Details are shown in Table~\ref{tab:opposite_color_bishops}.
For Figure~\ref{fig:opp_color_bishop_example_2}, an all-zero input is used as the baseline.
The neural network integrated with the FX-representation, shown in the representative position in Figure~\ref{fig:miller_saidy}, produces value estimates that tend to approach zero for drawn positions. On the other hand, the same network assigns significantly higher value scores for positions that represent decisive victories, as demonstrated in Figure~\ref{fig:kotov_botvinik}.

%% file: tables/opposite_color_bishops.tex
\definecolor{cold}{rgb}{0,0,1}
\definecolor{warm}{rgb}{1,0,0}

\ifarxiv
\begin{table*}[t]
\centering
\caption{Comparison of evaluations between the FX-network and AlphaZero for opposite color bishop endgames. The evaluations are presented based on the current player's perspective. The chess positions are sourced from \texttt{https://en.wikipedia.org/wiki/Opposite-colored\_bishops\_endgame} as of October 19, 2023. Ground truth evaluations indicate whether the position should result in a draw (DRAW) or a win for White (WHITE WIN) or Black (BLACK WIN). Net Eval and FX-Net Eval are the evaluations produced by AlphaZero's default and FX-network, respectively. Abs. Net Eval and Abs. FX-Net Eval indicate the respective absolute values. Values close to 0 are better for drawn positions, while values close to $-$1 or 1 are better for winning positions. The best evaluation in comparison to the absolute ground truth evaluation are denoted in \textbf{bold}.}
\label{tab:opposite_color_bishops}
    \resizebox{\textwidth}{!}{ % Adjust to fit the entire table to the column width
\begin{tabular}{
  l
  c
  S[table-format=1.4]
  S[table-format=1.4]
  S[table-format=1.4]
  S[table-format=1.4]
}
        \toprule
        \textbf{FEN} & \textbf{Ground Truth} & \multicolumn{1}{l}{\textbf{Net Eval}} & \multicolumn{1}{l}{\textbf{Abs. Net Eval}} & \multicolumn{1}{l}{\textbf{FX-Net Eval}} & \multicolumn{1}{l}{\textbf{Abs. FX-Net Eval}} \\
        \midrule
        8/2k1b3/2P5/3KP2B/8/8/8/8 w - - 0 56 & DRAW & 0.3332 & 0.3332 & \textbf{0.2025} & \textbf{0.2025} \\
        8/3k4/8/2pK4/8/4b1p1/8/5B2 w - - 0 56 & DRAW & -0.4940 & 0.4940 & \textbf{-0.2304} & \textbf{0.2304} \\
        5k2/8/8/7p/1b1p4/8/B7/5K2 b - - 0 56 & DRAW & 0.4385 & 0.4385 & \textbf{0.2500} & \textbf{0.2500} \\
        8/2b1k3/8/1B1PP3/3K4/8/8/8 w - - 0 56 & DRAW & \textbf{0.4313} &\textbf{ 0.4313} & 0.4561 & 0.4561 \\
        8/2k5/4Bp2/2b1p1p1/4K2p/7P/8/8 b - - 0 56 & DRAW & \textbf{0.1648} & \textbf{0.1648} & 0.2121 & 0.2121 \\
        8/8/8/5B2/1p3b2/2k1p3/8/5K2 w - - 0 56 & DRAW & -0.6412 & 0.6412 & \textbf{-0.4157} & \textbf{0.4157} \\
        8/3k4/p2P4/2P4p/2bB4/P6P/5K2/8 w - - 0 56 & DRAW & \textbf{0.3874} & \textbf{0.3874} & 0.4465 & 0.4465 \\
        7b/4k2P/6K1/2p2P2/7P/1B6/8/8 b - - 0 56 & DRAW & -0.6286 & 0.6286 & \textbf{-0.6233} & \textbf{0.6233} \\
        4k2b/7P/5PK1/7P/8/1B6/8/8 w - - 0 56 & DRAW & \textbf{0.7649} & \textbf{0.7649} & 0.8562 & 0.8562 \\
        8/5pK1/4k3/6B1/5PbP/6P1/8/8 b - - 0 56 & DRAW & -0.4810 & 0.4810 & \textbf{-0.4294} & \textbf{0.4294} \\
        2r3k1/5ppp/p7/5q2/3P4/b2B2P1/P1R2P1P/5QK1 b - - 0 56 & DRAW & \textbf{-0.5099} & \textbf{0.5099} & -0.5594 & 0.5594 \\
        5k2/5pp1/p6p/5B2/3P4/6P1/P3KP1P/2b5 w - - 0 56 & DRAW & \textbf{0.3222} & \textbf{0.3222} & 0.4718 & 0.4718 \\
        3b4/p4B1p/8/6k1/6P1/8/1P3PK1/8 w - - 0 56 & DRAW & 0.2920 & 0.2920 & \textbf{0.0783} & \textbf{0.0783} \\
        6B1/4b3/7p/3Pk2P/6PP/7K/8/8 w - - 0 56 & DRAW & \textbf{0.4867} & \textbf{0.4867} & 0.5590 & 0.5590 \\
        8/8/8/7p/2p5/5K1k/2Bb4/8 w - - 0 56 & DRAW & -0.3318 & 0.3318 & \textbf{-0.1916} & \textbf{0.1916} \\
        3R4/4BK1k/r5p1/2P2bP1/8/8/8/8 w - - 0 56 & WHITE WIN & 0.8304 & 0.8304 & \textbf{0.8714} & \textbf{0.8714} \\
        8/2k1b3/2P5/3K1P1B/8/8/8/8 w - - 0 56 & WHITE WIN & \textbf{0.4072} & \textbf{0.4072} & 0.3321 & 0.3321 \\
        3k1b2/8/3PP3/1B1K4/8/8/8/8 w - - 0 56 & WHITE WIN & 0.7121 & 0.7121 & \textbf{0.8229} & \textbf{0.8229} \\
        8/2k5/2P1K3/6p1/5p2/2b2B1P/6P1/8 b - - 0 56 & WHITE WIN & \textbf{-0.2412} & \textbf{0.2412} & -0.1441 & 0.1441 \\
        8/8/4b1p1/2Bp3p/5P1P/1pK1Pk2/8/8 b - - 0 56 & BLACK WIN & 0.4053 & 0.4053 & \textbf{0.4225} & \textbf{0.4225} \\
        \midrule
        Mean values for draws $\downarrow$ & 0 &  & 0.4472 &  & \textbf{0.3988} \\
        Mean values for wins $\uparrow$ & 1 &  & \textbf{0.5192} &  & 0.5186 \\
        \bottomrule
\end{tabular}
}
\end{table*}
\else
\begin{table*}[t]
\centering
\caption{Comparison of evaluations between the FX-network and AlphaZero for opposite color bishop endgames. The evaluations are presented based on the current player's perspective. The chess positions are sourced from \texttt{https://en.wikipedia.org/wiki/Opposite-colored\_bishops\_endgame} as of October 19, 2023. Ground truth evaluations indicate whether the position should result in a draw (DRAW) or a win for White (WHITE WIN) or Black (BLACK WIN). Net Eval and FX-Net Eval are the evaluations produced by AlphaZero's default and FX-network, respectively. Abs. Net Eval and Abs. FX-Net Eval indicate the respective absolute values. Values close to 0 are better for drawn positions, while values close to $-$1 or 1 are better for winning positions. The best evaluation in comparison to the absolute ground truth evaluation are denoted in \textbf{bold}.}
\label{tab:opposite_color_bishops}
\begin{tabular}{
  l
  c
  S[table-format=1.4]
  S[table-format=1.4]
  S[table-format=1.4]
  S[table-format=1.4]
}
        \toprule
        \textbf{FEN} & \textbf{Ground Truth} & \multicolumn{1}{l}{\textbf{Net Eval}} & \multicolumn{1}{l}{\textbf{Abs. Net Eval}} & \multicolumn{1}{l}{\textbf{FX-Net Eval}} & \multicolumn{1}{l}{\textbf{Abs. FX-Net Eval}} \\
        \midrule
        8/2k1b3/2P5/3KP2B/8/8/8/8 w - - 0 56 & DRAW & 0.3332 & 0.3332 & \textbf{0.2025} & \textbf{0.2025} \\
        8/3k4/8/2pK4/8/4b1p1/8/5B2 w - - 0 56 & DRAW & -0.4940 & 0.4940 & \textbf{-0.2304} & \textbf{0.2304} \\
        5k2/8/8/7p/1b1p4/8/B7/5K2 b - - 0 56 & DRAW & 0.4385 & 0.4385 & \textbf{0.2500} & \textbf{0.2500} \\
        8/2b1k3/8/1B1PP3/3K4/8/8/8 w - - 0 56 & DRAW & \textbf{0.4313} &\textbf{ 0.4313} & 0.4561 & 0.4561 \\
        8/2k5/4Bp2/2b1p1p1/4K2p/7P/8/8 b - - 0 56 & DRAW & \textbf{0.1648} & \textbf{0.1648} & 0.2121 & 0.2121 \\
        8/8/8/5B2/1p3b2/2k1p3/8/5K2 w - - 0 56 & DRAW & -0.6412 & 0.6412 & \textbf{-0.4157} & \textbf{0.4157} \\
        8/3k4/p2P4/2P4p/2bB4/P6P/5K2/8 w - - 0 56 & DRAW & \textbf{0.3874} & \textbf{0.3874} & 0.4465 & 0.4465 \\
        7b/4k2P/6K1/2p2P2/7P/1B6/8/8 b - - 0 56 & DRAW & -0.6286 & 0.6286 & \textbf{-0.6233} & \textbf{0.6233} \\
        4k2b/7P/5PK1/7P/8/1B6/8/8 w - - 0 56 & DRAW & \textbf{0.7649} & \textbf{0.7649} & 0.8562 & 0.8562 \\
        8/5pK1/4k3/6B1/5PbP/6P1/8/8 b - - 0 56 & DRAW & -0.4810 & 0.4810 & \textbf{-0.4294} & \textbf{0.4294} \\
        2r3k1/5ppp/p7/5q2/3P4/b2B2P1/P1R2P1P/5QK1 b - - 0 56 & DRAW & \textbf{-0.5099} & \textbf{0.5099} & -0.5594 & 0.5594 \\
        5k2/5pp1/p6p/5B2/3P4/6P1/P3KP1P/2b5 w - - 0 56 & DRAW & \textbf{0.3222} & \textbf{0.3222} & 0.4718 & 0.4718 \\
        3b4/p4B1p/8/6k1/6P1/8/1P3PK1/8 w - - 0 56 & DRAW & 0.2920 & 0.2920 & \textbf{0.0783} & \textbf{0.0783} \\
        6B1/4b3/7p/3Pk2P/6PP/7K/8/8 w - - 0 56 & DRAW & \textbf{0.4867} & \textbf{0.4867} & 0.5590 & 0.5590 \\
        8/8/8/7p/2p5/5K1k/2Bb4/8 w - - 0 56 & DRAW & -0.3318 & 0.3318 & \textbf{-0.1916} & \textbf{0.1916} \\
        3R4/4BK1k/r5p1/2P2bP1/8/8/8/8 w - - 0 56 & WHITE WIN & 0.8304 & 0.8304 & \textbf{0.8714} & \textbf{0.8714} \\
        8/2k1b3/2P5/3K1P1B/8/8/8/8 w - - 0 56 & WHITE WIN & \textbf{0.4072} & \textbf{0.4072} & 0.3321 & 0.3321 \\
        3k1b2/8/3PP3/1B1K4/8/8/8/8 w - - 0 56 & WHITE WIN & 0.7121 & 0.7121 & \textbf{0.8229} & \textbf{0.8229} \\
        8/2k5/2P1K3/6p1/5p2/2b2B1P/6P1/8 b - - 0 56 & WHITE WIN & \textbf{-0.2412} & \textbf{0.2412} & -0.1441 & 0.1441 \\
        8/8/4b1p1/2Bp3p/5P1P/1pK1Pk2/8/8 b - - 0 56 & BLACK WIN & 0.4053 & 0.4053 & \textbf{0.4225} & \textbf{0.4225} \\
        \midrule
        Mean values for draws $\downarrow$ & 0 &  & 0.4472 &  & \textbf{0.3988} \\
        Mean values for wins $\uparrow$ & 1 &  & \textbf{0.5192} &  & 0.5186 \\
        \bottomrule
\end{tabular}
\end{table*}
\fi
% data can be found here:
% https://docs.google.com/spreadsheets/d/1JfKJ_9V17a0pbe2a9eGhr4hQAV7xWCunvgonYRIJIE8/edit#gid=0
% and can be converted here;
% https://www.tablesgenerator.com/#